%% file: main.tex
\documentclass[10pt,twocolumn,letterpaper]{article}

\usepackage[pagenumbers]{iccv} 

\input{sec/imports}

\newcommand{\edit}[1]{#1}

\def\paperID{13523} 
\def\confName{ICCV}
\def\confYear{2025}

\title{
Detection, Pose Estimation and Segmentation for Multiple Bodies:\\
Closing the Virtuous Circle}
\input{sec/authors}

\begin{document}
\twocolumn[{%
\renewcommand\twocolumn[1][]{#1}%
\maketitle
\vspace{-1.0cm}
\begin{center}
    \centering
    \captionsetup{type=figure}
    \begin{subfigure}{0.150\linewidth}
        \centering
        \includegraphics[width=\textwidth]{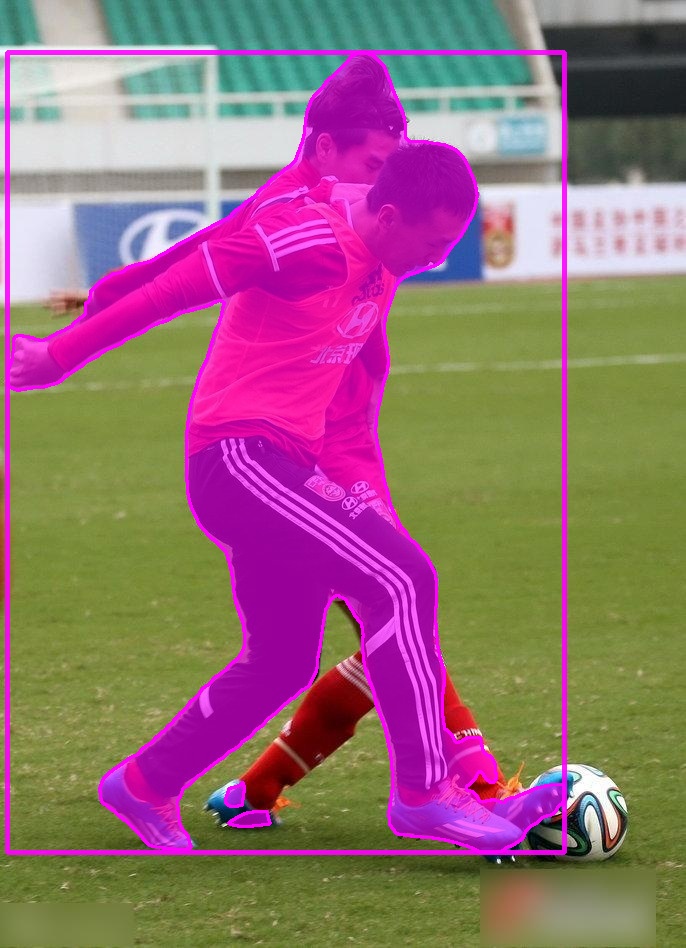}
        \caption*{
        (A1) Detection
        }
    \end{subfigure}
    \begin{subfigure}{0.150\linewidth}
        \centering
        \includegraphics[width=\textwidth]{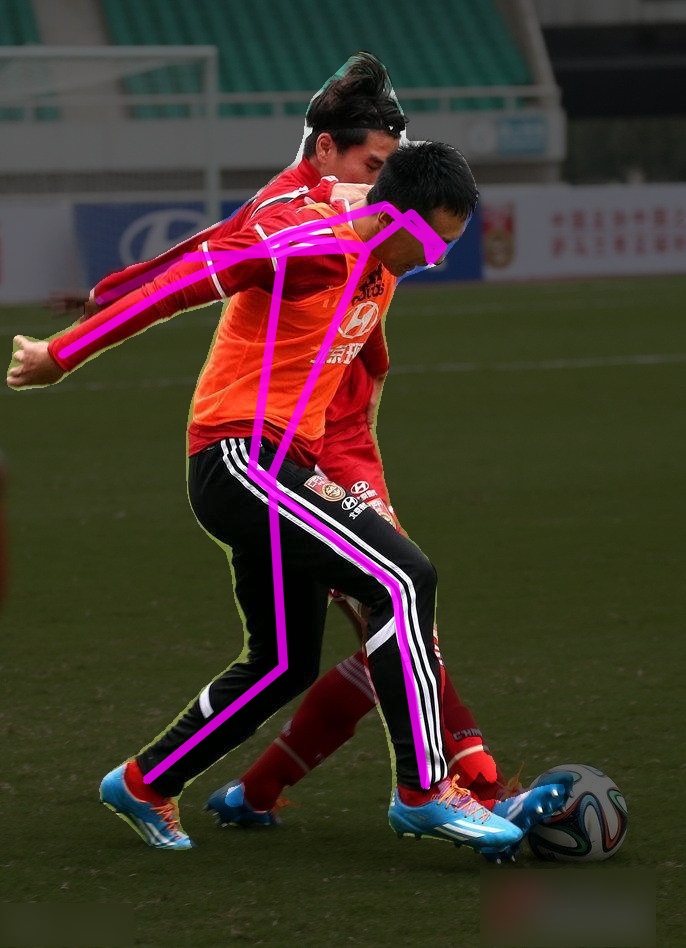}
        \caption*{
        (B1) Pose estimation
        }
    \end{subfigure}
    \begin{subfigure}{0.150\linewidth}
        \centering
        \includegraphics[width=\textwidth]{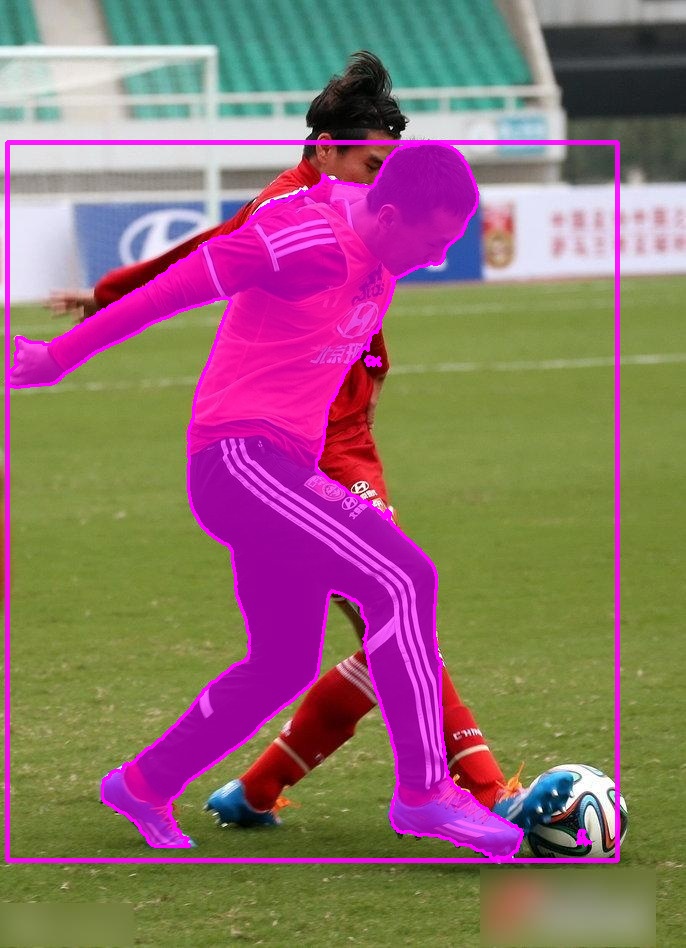}
        \caption*{
        (C1) Mask refinement
        }
    \end{subfigure}
    \begin{subfigure}{0.150\linewidth}
        \centering
        \includegraphics[width=\textwidth]{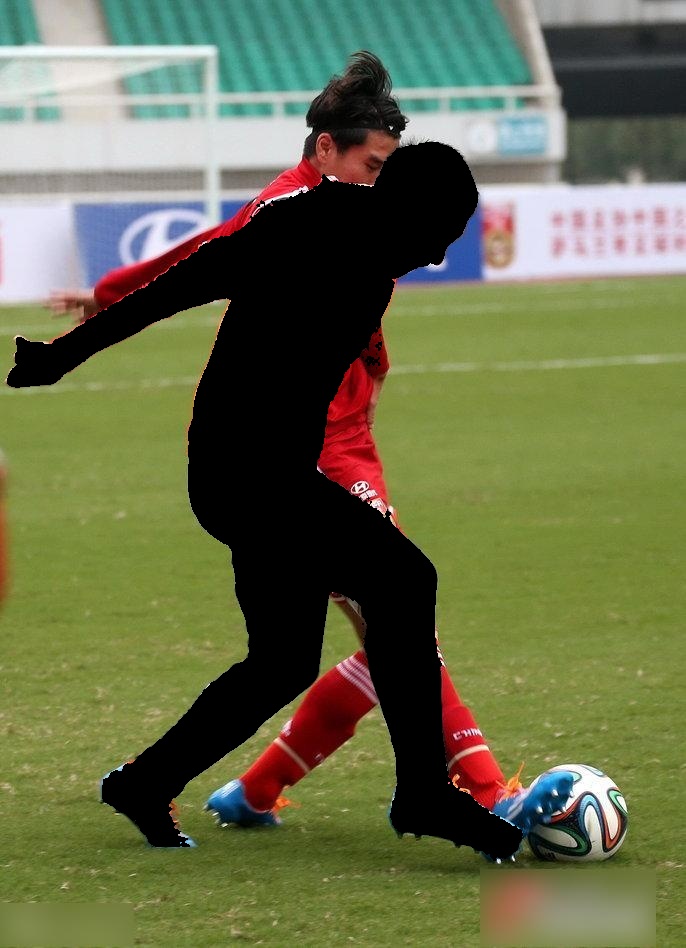}
        \caption*{
        (D1) Mask-out
        }
        \label{fig:BMP-visualization-maskout}
    \end{subfigure}
    \begin{subfigure}{0.150\linewidth}
        \centering
        \includegraphics[width=\textwidth]{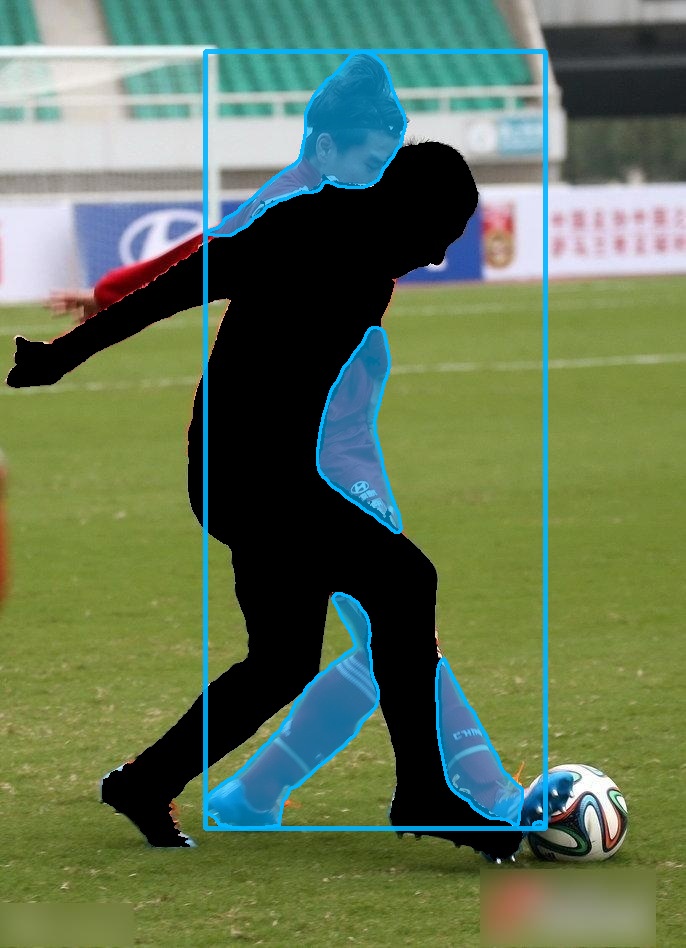}
        \caption*{
        (A2) Detection
        }
    \end{subfigure}
    \hspace{-0.8cm}
    \begin{subfigure}{0.150\linewidth}
        \centering
        \includegraphics[height=4cm,page=2]{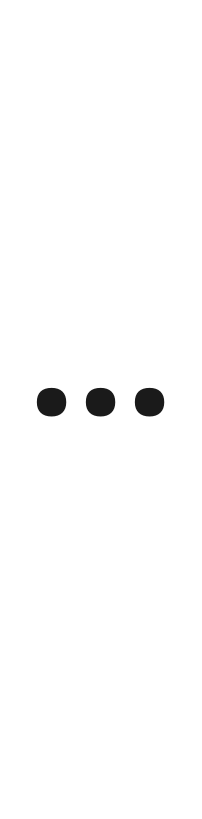}
    \end{subfigure}
    \hspace{-0.8cm}
    \begin{subfigure}{0.150\linewidth}
        \centering
        \includegraphics[width=\textwidth]{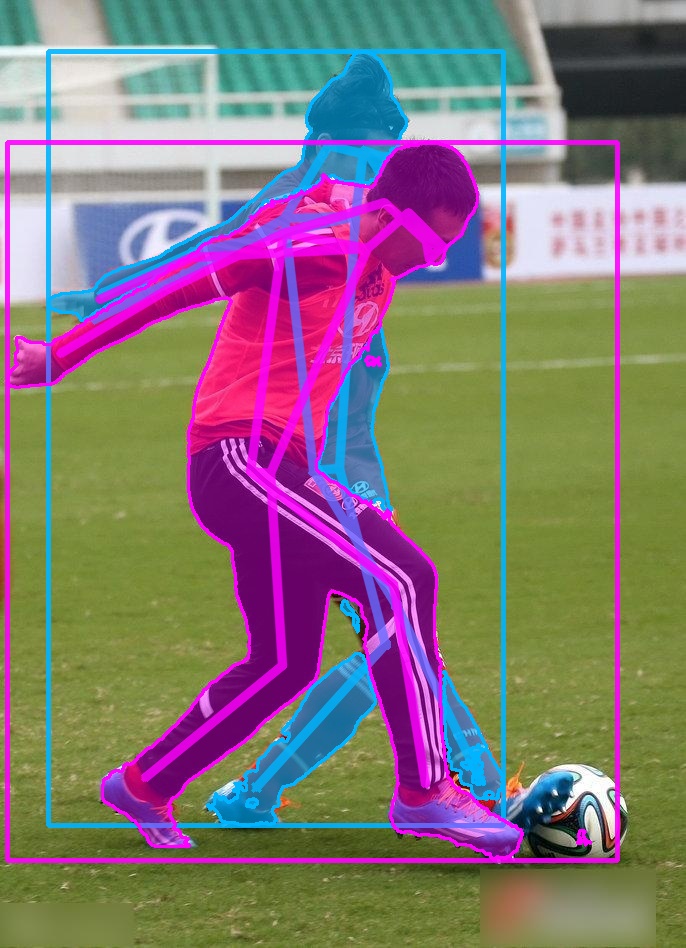}
        \caption*{
        \hspace{-7ex}Output: bboxes, poses, masks
        }
    \end{subfigure}
    \caption{
    \textbf{The BBox-Mask-Pose (BMP) method}.
    Steps (A) -- (D) repeat until no new detections found in step (A).
    Here, the \textcolor{BckPlayerBlue}{background player} is undetected in step (A1). 
    BMP correctly fits the \textcolor{ForePlayerMagenta}{foreground player's} pose (B1) which leads to correction of his segmentation and bbox (C1).
    After masking the \textcolor{ForePlayerMagenta}{foreground player} (D1), the \textcolor{BckPlayerBlue}{background player} is detected (A2), his body correctly segemented and pose estimated. Right: the BMP output. Note: the loop can start with a bounding box (A), pose (B), or segmentation mask~(C).
    }
    \label{fig:BMP-visualization}
\end{center}%
}]

\input{sec/paper}

\input{sec/acknowledgements}

{
    \small
    \bibliographystyle{ieeenat_fullname}
    \bibliography{main}
}

\input{sec/supplementary}

\end{document}

%% file: sec/imports.tex
\usepackage{graphicx}
\usepackage{amsmath}
\usepackage{amssymb}
\usepackage{booktabs}
\usepackage{enumitem}
\usepackage{fancyhdr}

\usepackage{amsthm}
\usepackage{pifont}
\usepackage[dvipsnames]{xcolor}
\usepackage{colortbl}
\usepackage{pgfplots}
\pgfplotsset{compat=1.18}
\usepackage{subcaption}
\usepackage{multirow}
\usepackage{tablefootnote}
\usepackage{relsize}
\usepackage[accsupp]{axessibility} 

\usepackage[linesnumbered,commentsnumbered,ruled,lined]{algorithm2e}

\usepackage[normalem]{ulem}
\usepackage[leftcaption]{sidecap} 

\definecolor{cvprblue}{rgb}{0.21,0.49,0.74}
\usepackage[pagebackref,breaklinks,colorlinks,allcolors=cvprblue]{hyperref}

\usepackage[capitalize]{cleveref}
\crefname{section}{Sec.}{Secs.}
\Crefname{section}{Section}{Sections}
\Crefname{table}{Table}{Tables}
\crefname{table}{Tab.}{Tabs.}
\crefname{algorithm}{Alg.}{Algs.}
\Crefname{algorithm}{Algorithm}{Algorithms}

\newcommand{\cmark}{\ding{51}}%
\newcommand{\xmark}{\ding{55}}%

\definecolor{LightCyan}{rgb}{0.7,1,1}
\definecolor{LightYellow}{rgb}{1,1,0.7}
\definecolor{LightOrange}{rgb}{1,0.8,0.5}
\definecolor{DarkGreen}{rgb}{0.0,0.5,0.0}
\definecolor{Burgundy}{rgb}{0.5, 0, 0.125}
\definecolor{HunterGreen}{rgb}{0.207, 0.367, 0.23}
\definecolor{PastelGreen}{rgb}{0.9, 1.0, 0.9} 
\definecolor{LightBlue}{RGB}{220, 245, 255}
\definecolor{DarkGray}{RGB}{120, 120, 120}
\definecolor{BckPlayerBlue}{RGB}{0, 180, 255}
\definecolor{ForePlayerMagenta}{RGB}{255, 0, 255}
\definecolor{BrightBlue}{RGB}{0, 150, 255}
\definecolor{CobaltBlue}{RGB}{0, 71, 17}
\definecolor{BlueGreen}{RGB}{8, 143, 143}

\newcolumntype{A}{>{\columncolor{LightCyan}}r}
\newcolumntype{B}{>{\columncolor{LightYellow}}r}
\newcolumntype{D}{>{\columncolor{LightOrange}}r}

\newcommand{\grn}[1]{\textcolor{DarkGreen}{#1}}
\newcommand{\red}[1]{\textcolor{Burgundy}{#1}}

\newcommand{\grnb}[1]{\textcolor{DarkGreen}{\textbf{#1}}}
\newcommand{\redb}[1]{\textcolor{Burgundy}{\textbf{#1}}}

\newcommand{\impr}[1]{\footnotesize{\textcolor{DarkGreen}{#1}}}
\newcommand{\decr}[1]{\footnotesize{\textcolor{red}{#1}}}

\newcommand{\cplus}{\bigcirc\hspace{-0.89em}+}
\newcommand{\cminus}{\bigcirc\hspace{-0.89em}-}

%% file: sec/authors.tex
\author{
\vspace{-0.8cm}\\
Miroslav Purkrabek \hspace{0.05cm} and \hspace{0.05cm} Jiri Matas
\vspace{0.1cm}\\
Visual Recognition Group\\
Czech Technical University in Prague\\
{\tt\small purkrmir@fel.cvut.cz}
}


%% file: sec/paper.tex
\begin{abstract}
\vspace{-1.5em}

\noindent
Human pose estimation methods work well on isolated people but struggle with multiple-bodies-in-proximity scenarios.
Previous work has addressed this problem by conditioning pose estimation by detected bounding boxes or keypoints, but overlooked instance masks.
We propose to iteratively enforce mutual consistency of bounding boxes, instance masks, and poses.
The introduced BBox-Mask-Pose (BMP) method uses three specialized models that improve each other's output in a closed loop.
All models are adapted for mutual conditioning, which improves robustness in multi-body scenes.
MaskPose, a new mask-conditioned pose estimation model, is the best among top-down approaches on OCHuman. 
BBox-Mask-Pose pushes SOTA on OCHuman dataset in all three tasks -- detection, instance segmentation, and pose estimation.
It also achieves SOTA performance on COCO pose estimation. 
The method is especially good in scenes with large instances overlap, where it improves detection by 39\% over the baseline detector. 
With small specialized models and faster runtime, BMP is an effective alternative to large human-centered foundational models. 
Code and models are available on the \href{https://mirapurkrabek.github.io/BBox-Mask-Pose/}{project website} \footnote{\url{MiraPurkrabek.github.io/BBox-Mask-Pose/}}.
%
%
%
%
\end{abstract}
\vspace{-0.5em}

\section{Introduction}
\label{sec:intro}

Human pose estimation (HPE) plays a crucial role in tasks such as action detection and gesture recognition. 
It is a challenging problem, especially in multi-body scenes where people overlap, leading to issues such as merged bounding boxes or collapsed poses. Results on multi-body datasets
are far from saturated, with state-of-the-art below 50\% \cite{BUCTD}.



HPE approaches differ in how they use \textit{conditioning} to guide predictions.
Top-down methods \cite{ViTPose,MIPNet} are conditioned by bounding boxes, estimating poses within image crops defined by a detector, while
single-stage and bottom-up methods \cite{CID, OpenPose} are not conditioned at all. 
Pose-refining methods, such as BUCTD \cite{BUCTD}, introduce conditioning by prior pose estimates, iteratively refining predictions to improve accuracy.

Bounding boxes, masks, and poses represent different aspects of the human body, at different levels of granularity. 
Bounding boxes are easy to annotate and effective for detecting small instances but lack detail in crowded scenes.
Segmentation masks are more detailed, but are costly to annotate and less common than bboxes.
Poses provide anatomical detail, but are less effective for direct detection. 
Detectors, segmentators, and pose estimators are often trained on different datasets, and their combination increases variance in training data.

The proposed BBox-Mask-Pose (BMP) method extends conditioning to masks and integrates bboxes, masks, and poses into feedback loop (\cref{fig:BMP-visualization}).
BMP uses three specialized models that iteratively refine each other's output, allowing (re-)detection, segmentation, and pose estimation to achieve consistent results and performance gains, especially in multi-body scenarios.
Specifically, the models are:
\begin{itemize}
    \item Fine-tuned RTMDet \cite{RTMDet}: A detector that could be conditioned by segmentation masks and ignores masked-out instances.
    
    \item MaskPose: ViTPose-based \cite{ViTPose} pose estimation model conditioned by instance segmentation masks and bounding boxes. Its pose estimation is more robust in dense scenes than the previous top-down SOTA.
    
    \item SAM2 (Segment Anything Model) \cite{SAM2}, conditioned by suitably selected pose keypoints, which improves segmentation capabilities and facilitates information passing between bounding box locations and pose estimates.
\end{itemize}

For pose estimation, BMP sets the new state-of-the-art performance on OCHuman, while also achieving SOTA performance on the COCO dataset.
For detection and instance segmentation, BMP sets the new SOTA on OCHuman.
None of the models in the loop were trained on OCHuman data and the same hyper parameters are used for evaluation on standard dataset (COCO) and multi-body scenes (OCHuman). 

Ablations show that mutual conditioning creates a cycle that improves the accuracy of all components.
The combination of an object detector with a model that ``understands'' the object structure could generalize to tasks where specialized models interpret the structure, as HPE models do for human anatomy.
Moderately sized models (RTMDet-L \cite{RTMDet}, ViTPose-B \cite{ViTPose}, SAM-B+ \cite{SAM2}) are used in all experiments.
The modular structure of BMP allows any component to be replaced by a larger or superior alternative to achieve improved performance.

In summary, \textbf{the main contribution} is the BBox-Mask-Pose loop, a new method for robust detection, segmentation and pose estimation in multi-body scenes. 
The core idea of BMP is to enforce mutual consistency between different representations of a human body \edit{through mutual conditioning}.
Experiments show that three specialized models are an effective alternative to data- and computationally expensive \edit{shared-features} foundational models.

Other \textbf{technical contributions} are MaskPose, the first top-down HPE model conditioned by detected masks, the fine-tuned detector ignoring masked-out instances, and the keypoint selection algorithm for automated SAM prompting for pose-to-mask estimation.

\section{Related work}
\label{sec:related}

\textbf{Datasets.}
There are various datasets for 2D human pose estimation. Most notable are: COCO \cite{COCO}, MPII \cite{MPII} and AIC \cite{AIC}.
Datasets like OCHuman \cite{pose2seg} and CrowdPose \cite{CrowdPose} focus on multibody problems such as occlusion and self-occlusion.
OCHuman is too small for large-scale training and is traditionally used only for evaluation.
CrowdPose is big enough for training but is unsuitable for evaluation in multi-dataset setup as it mixes train and test sets of COCO, MPII and AIC.
For COCO and related datasets, the evaluation metric is Object Keypoint Similarity (OKS), while Percentage of Correct Keypoints (PCKh) is used for MPII.
In addition to the pose estimation, other datasets focus on human crowds, e.g. CrowdHuman \cite{CrowdHuman} on detection \edit{or CIHP \cite{CIHP} on human parsing}.

\textbf{Human pose estimation.}
There are two main approaches to 2D human pose estimation: top-down and detector-free.
Detector-free can be further divided into single-stage \cite{POET,PETR,AdaptivePose++,CID}, bottom-up \cite{DEKR,OpenPose,BottomUpSeg} and hybrid \cite{BUCTD}.

Top-down methods \cite{ViTPose,HRNet,SWIN,HRFormer,MIPNet} use person detector to detect bounding boxes and estimate one skeleton for each bounding box.
They leverage big progress in human detection and specialize on understanding of human structure.
Top-down methods are the most successful on datasets such as COCO, MPII or AIC but struggle on crowded datasets (OCHuman) due to low-quality detections.
Most notably, ViTPose \cite{ViTPose} combines multiple datasets into one strong backbone and sets a strong baseline, setting up state-of-the-art performance on most datasets.
While conditioning pose estimation on bounding boxes (bbox-to-pose; standard top-down approach) is well researched, conditioning pose on masks (mask-to-pose) was not explored.

On the other hand, detector-free models do not achieve SOTA performance on COCO but are superior to top-down methods on OCHuman as they are specialized on decoupling close-interaction instances.
The most successful model, BUCTD \cite{BUCTD}, conditions top-down pose estimation by previously estimated keypoints from bottom-up methods.
It is a pose-refinement method which has state-of-the-art results on OCHuman datasets due to its strong ability to decouple people close interactions.

\textbf{Foundational models.}
The latest directions in human body modeling are foundational models \cite{Sapiens,HULK,UniHCP,DeepSortLab}.
They learn general features describing human body that could be used for all human-related tasks such as segmentation, pose estimation etc.
Most notably, Sapiens 2b \cite{Sapiens} was trained on staggering 2M images and with 2B parameters is almost four times bigger than ViTPose-h.
Even with this size, foundational models perform comparatively or worse than much smaller specialized models. 

\textbf{Detectors.}
Object (or person) detection is one of the most researched problems in computer vision.
Huge models such as InternImage \cite{InternImage} or Co-DETR \cite{coDETR} holds SOTA performance on multiple datasets.
In our comparison, we use smaller almost real-time models RTMDet \cite{RTMDet}, ConvNeXt \cite{ConvNeXt} and HRNet \cite{HRNet} which have slightly lower performance but run much faster. 
In top-down HPE methods, the detector guides pose estimation, but the information never goes back to the detector.
\edit{
The topic of detection of objects in the image given a set of already detected objects has only few works.
Our detector masks out detected instances in the image and detects new instances.
The closest work is IterDet \cite{IterDet}, which also detects instances itteratively but masks detected instances in the feature space rather than in the image.
Other methods solve detection in crowds with advanced non-maxima suppression (NMS) techniques.
For example, PoseNMS \cite{PoseNMS} uses a human pose for non-maxima suppression.
}

\textbf{Segmentators.}
The idea of segmentation conditioned by human pose is not new.
Many models \cite{pose2seg,pose2instance,PosePlusSeg,PoSeg,MultiPoseSeg} estimate instance segmentation from either ground truth pose or estimated keypoints.
Other methods such as \cite{PoseSegTTA} use pose for test-time adaptation in instance segmentation.
The latest segmentation foundational model SAM2 \cite{SAM2} is conditioned not only by human pose but by any point(s).
Similarly to detection, conditioning mask by pose is well researched, but the other direction (conditioning pose by mask) remains unexplored.

\section{Method}
\label{sec:method}

\begin{figure}[tb]
    \centering
    \begin{subfigure}{0.93\linewidth}
        \centering
        \includegraphics[width=0.49\linewidth]{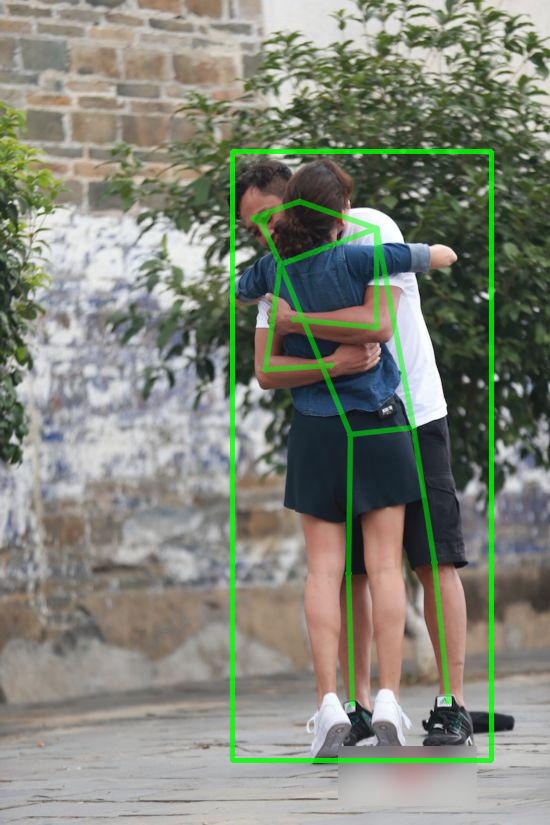}
        \hfill
        \includegraphics[width=0.49\linewidth]{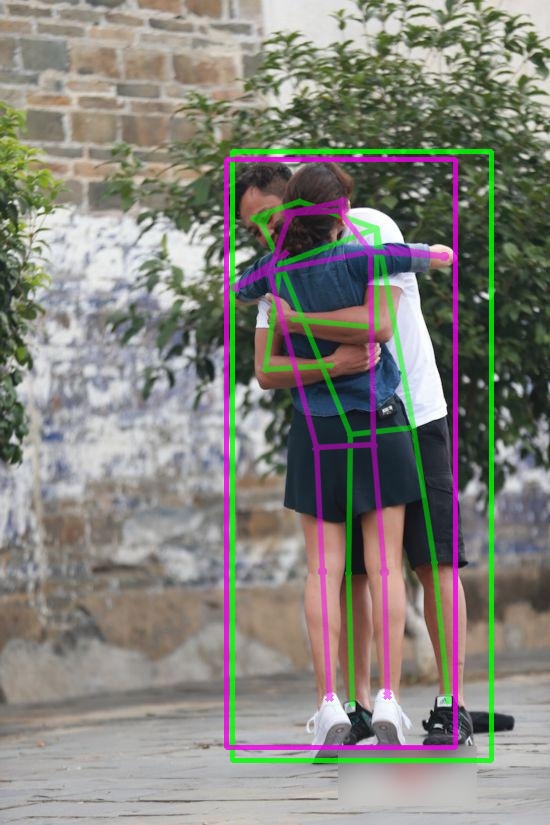}
        \caption{\textbf{A missed instance} which is detected in the second BMP iteration. Left -- RTMDet \cite{RTMDet} + MaskPose, right -- BMP.}
        \label{fig:examples-solved-2g1d_miss}
    \end{subfigure}
\\[1em]
    \begin{subfigure}{0.93\linewidth}
        \centering
        \includegraphics[width=0.49\textwidth]{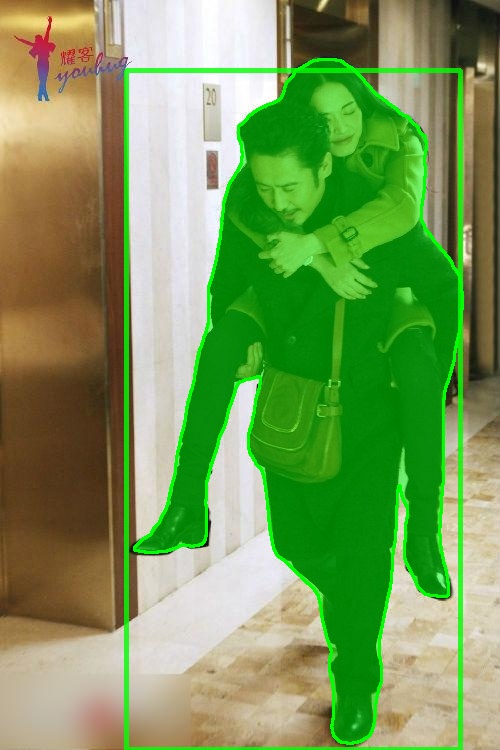}
        \hfill
        \includegraphics[width=0.49\textwidth]{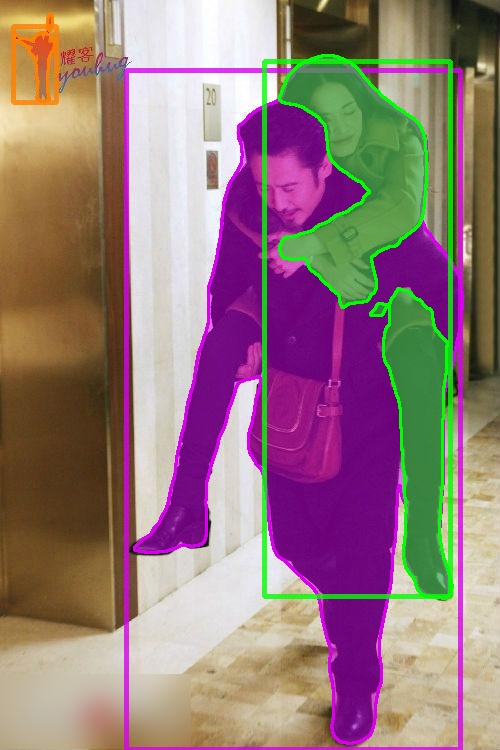}
        \caption{
        \textbf{Two instances in one detection} are resolved by refining segmentation masks with SAM \cite{SAM2} prompted by the detected pose.\\
        Left -- RTMDet \cite{RTMDet}, right -- BMP.
        Note that the detection of the woman is improved, but her right leg is still wrongly segmented.
        }
        \label{fig:examples-solved-2g1d_merge}
    \end{subfigure}
\\[1em]
    \begin{subfigure}{0.93\linewidth}
        \centering
        \includegraphics[width=0.49\textwidth]{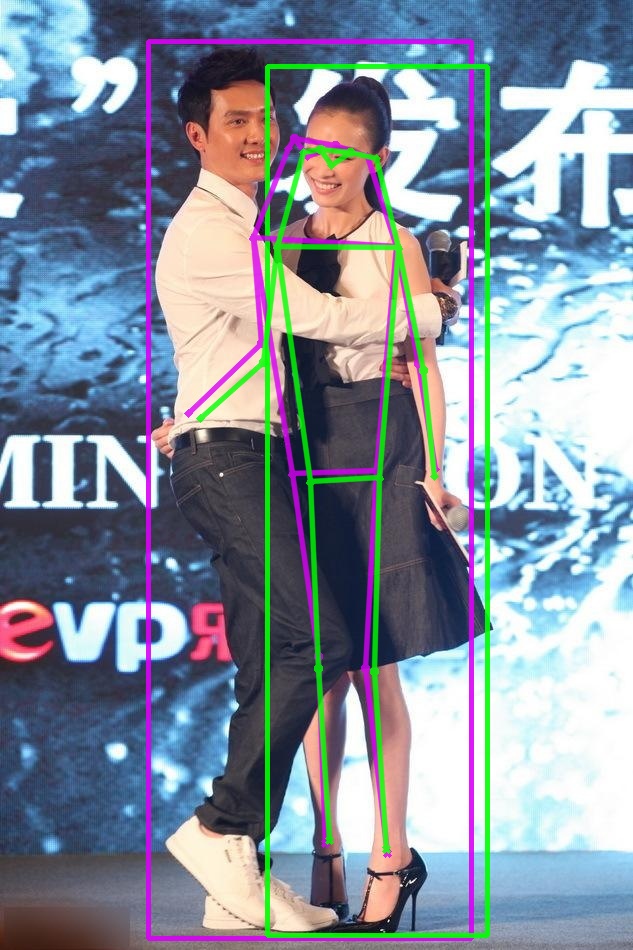}
        \hfill
        \includegraphics[width=0.49\textwidth]{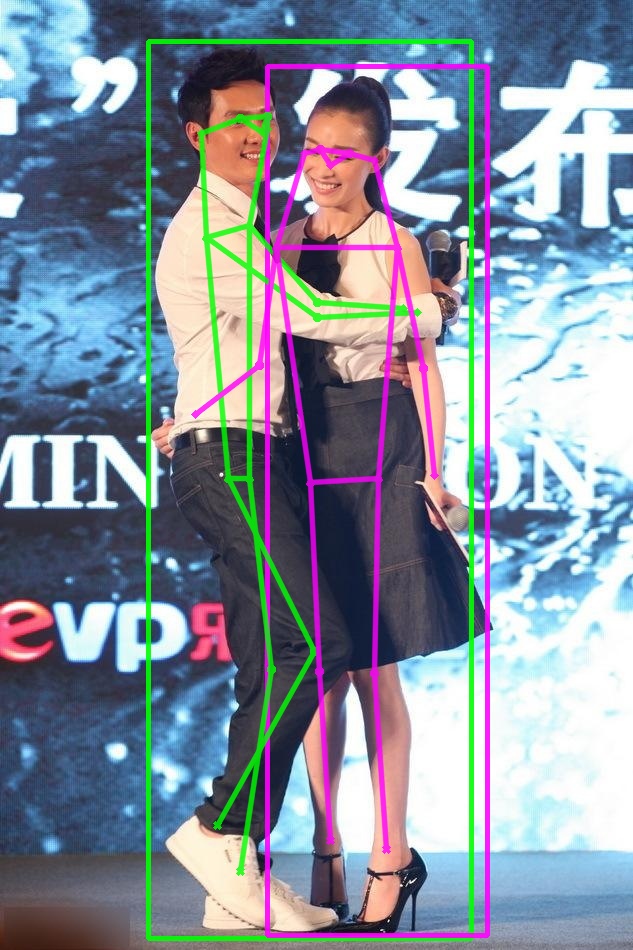}
        \caption{
        \textbf{Collapse of pose estimates for two instances} with correctly detected overlapping bboxes onto one body.
        Left -- ViTPose-B \cite{ViTPose} conditioned by bounding box, right -- MaskPose-b conditioned by masks.
        }
        \label{fig:examples-solved-2g2d}
    \end{subfigure}
    
    \caption{
    \textbf{BMP resolves detection errors}
    (top and middle) and pose errors (bottom) on OCHuman.
    Quantitative results in \cref{tab:det-ablation-study}.
    \vspace{-1.1cm}
    }
    \label{fig:examples-solved}
\end{figure}

The following sections detail the components of the BBox-Mask-Pose (BMP) method.
To create an iterative process that involves detection (bboxes $\mathcal{B}_i$), segmentation (binary masks $\mathcal{M}_i$) and pose estimation (keypoints $\mathcal{K}_i$), each component must be conditioned by the others.
We adapt the detector ($\mathcal{D}$) and pose estimator ($\mathcal{P}$) for mask conditioning and use Segment Anything Model 2 \cite{SAM2} ($\mathcal{S}$) to condition masks with bounding boxes and keypoints.

The BMP loop starts with the detector. It could start from any of the three representations, \edit{but bboxes are the easiest to obtain without any other input than the image}.

\subsection{Detection}
\label{sec:Method-detection}


The detector $\mathcal{D}$ detects bboxes $\mathcal{B}_i$ and masks $\mathcal{M}_i$ in the image $\mathcal{I}$ (\cref{eq:conditioned_detection}).
The image is masked by previously detected instances $\mathcal{M}_i$ as shown in \cref{fig:BMP-visualization} (A2).
\begin{align}
    \big( \mathcal{B}_i, \mathcal{M}_i \big) &= \mathcal{D} \big( \mathcal{I} \! \odot \! (1 - \bigcup_{i} \mathcal{M}_i  ) \big)
    \label{eq:conditioned_detection} \\
    \big( \mathcal{B}_i, \mathcal{M}_i \big) &= \mathcal{D} ^{init} \big( \mathcal{I} \big),
    \label{eq:pure_detection}
\end{align}
where $\odot$ stands for the Hadamard product of two matrices.

During masking, all pixels that belong to at least one mask $\mathcal{M}_i$ are set to black.
In the initial stage, there are no detected instances $\mathcal{M}_i$ and the detection phase becomes \cref{eq:pure_detection} -- the standard object detection task (\cref{fig:BMP-visualization} (A1)). 

Any standard detector could be used as $\mathcal{D}^{init}$.
For detector $\mathcal{D}$ conditioned by $\mathcal{M}_i$, we fine-tuned RTMDet~\cite{RTMDet} with \textit{instance removal} data augmentation simulating masked-out instances as in \cref{fig:BMP-visualization} (D1).
During training, randomly selected instances in the image are masked out and the model is trained not to predict them.
The fine-tuned detector retains its ability to detect instances in unmasked images, and the same model could be used for both $\mathcal{D}$ and $\mathcal{D}^{init}$. 

The masked pixels are set to non-transparent black.
When the mask is incorrect, non-transparent masking-out leads to information loss.
In the next section, MaskPose uses semi-transparent masking to improve robustness to incorrectly estimated masks.
Training the detector with semi-transparent masking led to much worse performance as the detector kept detecting masked-out instances.

RTMDet estimates both bounding boxes and segmentation masks.
MaskPose leverages estimated masks to predict more accurate poses in the next section.
Alternatively, a bbox-conditioned pose estimator could be used with the detector that estimates only bboxes. 

\subsection{Pose Estimation}
\label{sec:method-pose}


Traditional top-down methods (\cref{eq:pose_ViTPose}) rely solely on bounding boxes, cropping an image patch centered on the bounding box.
If multiple people appear in the same crop, the model estimates the pose of the central person but often merges body parts from others into a single skeleton.
We introduce MaskPose (\cref{eq:pose_MaskPose}), which builds on ViTPose \cite{ViTPose} and adapts it to use segmentation masks for conditioning.
\vspace{-1em}\begin{align}
    \mathcal{K}_i &= \mathcal{P} \big( \mathcal{I}, \mathcal{B}_i \big)
    \label{eq:pose_ViTPose}\\
    \mathcal{K}_i &= \mathcal{P}_{\alpha} \big( \alpha \mathcal{I} + (1{-}\alpha)(\mathcal{I} \! \odot \! \mathcal{M}_i ), \mathcal{B}_i \big)
    \label{eq:pose_MaskPose}
\end{align}

In \cref{eq:pose_MaskPose}, pose estimator $\mathcal{P}_{\alpha}$ is trained to predict pose in semi-transparently masked image $\mathcal{I}\odot\mathcal{M}_i$.
Pixels not belonging to mask $\mathcal{M}_i$ are darkened as shown in \cref{fig:BMP-visualization} (B1).

The model $\mathcal{P}_{\alpha}$ needs to be re-trained for a given $\alpha$.
Fully masking the background ($\alpha{=}0$) causes loss of contextual information, impairing MaskPose's recovery from inaccurate masks.
No masking ($\alpha{=}1$) reverts to a traditional bounding-box-based approach (\cref{eq:pose_ViTPose}).
All preliminary experiments (see Suppl.) with $\alpha \in [0.2, 0.8]$ had the same performance and we settled with $\alpha = 0.8$.
To enhance robustness to inaccurate masks, we randomly deform ground truth masks during training, allowing the model to predict keypoints outside the mask.

ViTPose trained in multi-dataset setup generalizes well across datasets, leveraging the strength of the ViT \cite{ViT} backbone.
ViTPose uses specialized head for each dataset with shared backbone.
MaskPose is also trained on the COCO, MPII, and AIC datasets but has a single head for all datasets.
The head predicts all 22 keypoints defined across COCO, AIC, and MPII, resulting in negligible performance loss compared to using separate heads.
MaskPose can thus be evaluated directly on any dataset without switching heads.



MaskPose has approximately the same number of parameters as ViTPose, differing only in head architecture and preprocessing.
These small changes enable MaskPose to perform similarly on standard datasets (COCO, AIC, MPII) while improving performance in multi-body scenarios.
Mask conditioning adapts the top-down method for multi-body cases, allowing detailed instance specification in densely overlapping scenes.

\subsection{Segmentation}
\label{sec:Method-segmentation}

\begin{figure}[tb]
    \centering
    \begin{subfigure}{\linewidth}
        \centering
        \includegraphics[width=0.49\linewidth,page=18]{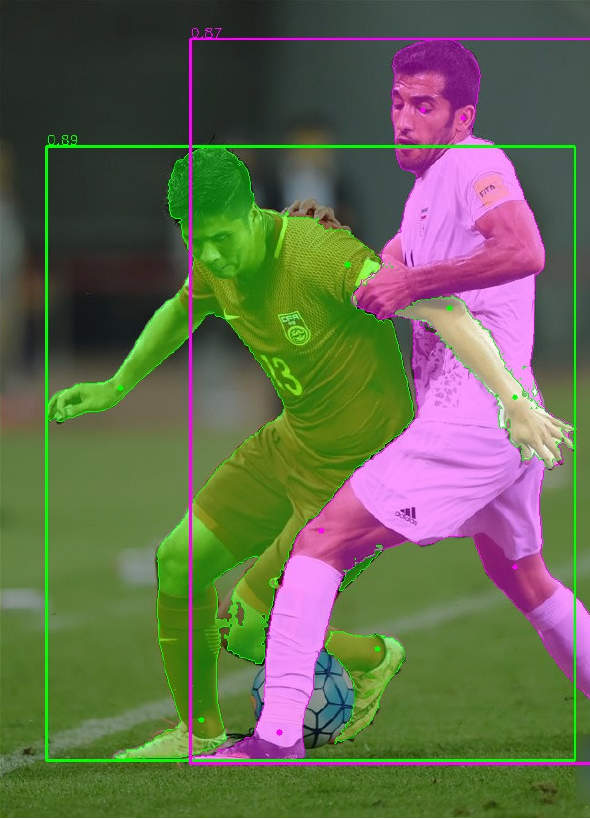}
        \hfill
        \includegraphics[width=0.49\linewidth,page=19]{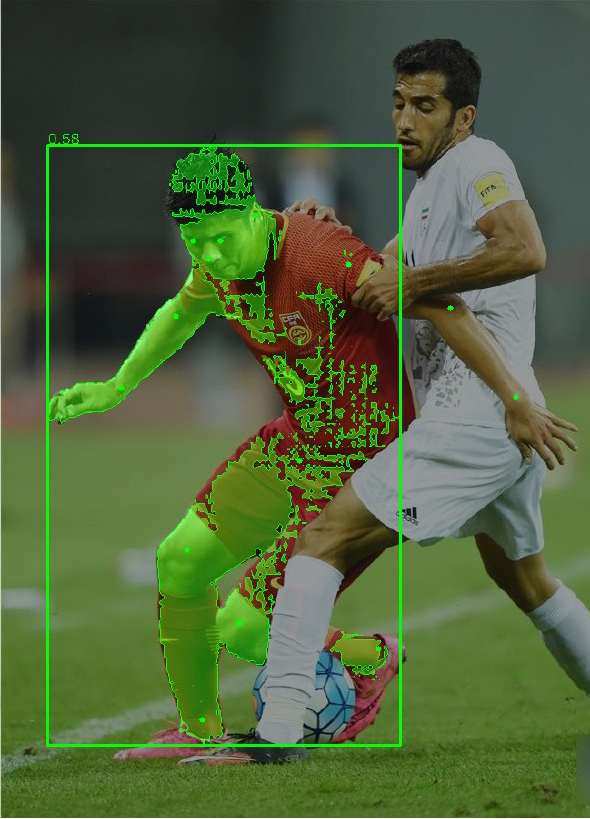}
        \caption{
        \textbf{The number of keypoints}.
        Too many points hinders performance.
        \\Left -- 6 keypoint prompts, right -- 13 correct prompts.
        }
        \label{fig:SAM-ablation-keypoints}
    \end{subfigure}
      \\[1em]  
    \begin{subfigure}{\linewidth}
        \centering
        \includegraphics[width=0.31\textwidth,page=21]{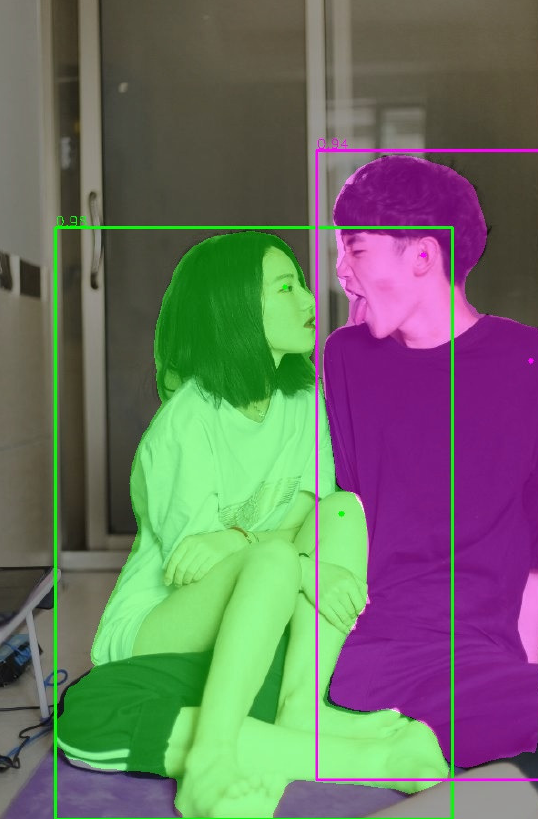}
        \hfill
        \includegraphics[width=0.31\textwidth,page=22]{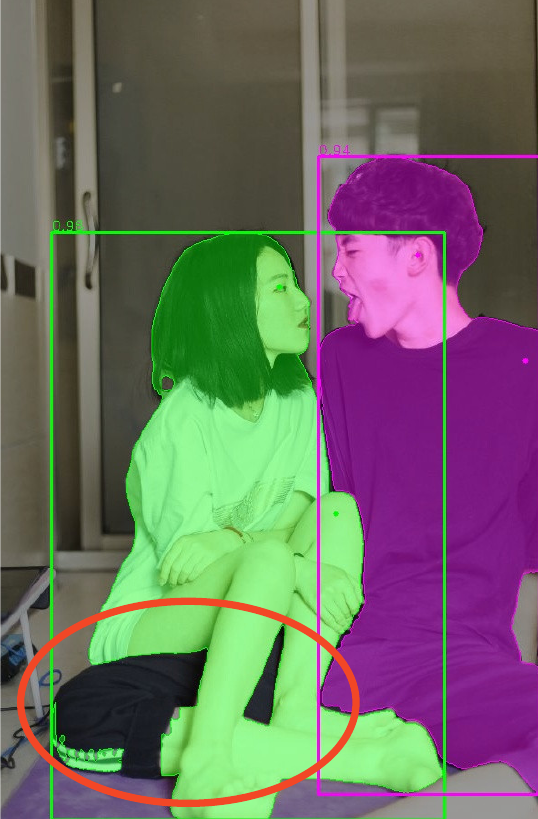}
        \hfill
      \includegraphics[width=0.31\textwidth,page=23]{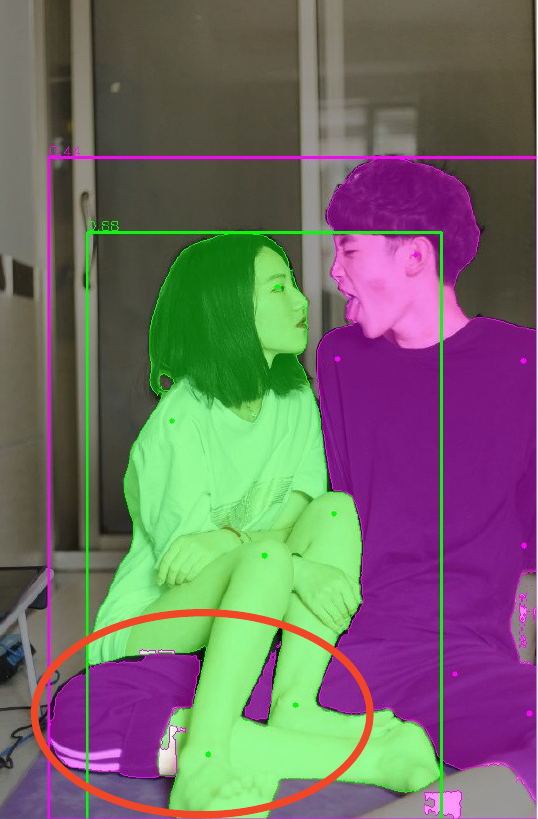}
        \caption{
        Prompting 
        \textbf{with and without a bounding box}. 
        Prompting with bbox prevents SAM from correcting body masks outside of the bounding box.
        \\Left -- RTMDet \cite{RTMDet},
        middle -- SAM with bbox, right -- SAM without bbox.
        }
        \label{fig:SAM-ablation-bbox}
    \end{subfigure}
    
    \caption{
    \textbf{SAM}: influence of prompting parameters.
    }
    \label{fig:SAM-ablation}
\end{figure}

We use Segment Anything Model v2 (SAM) \cite{SAM2} ($\mathcal{S}$) for mask generation, conditioned by estimated bounding boxes ($\mathcal{B}_i$) and keypoints ($\mathcal{K}_i$).
\vspace{-0.5em}\begin{align}
   \big( \mathcal{B}_i, \mathcal{M}_i \big) &= \mathcal{S} \big( \mathcal{I}, f(\mathcal{K}_i), g(\mathcal{B}_i)  \big)
\end{align}\vspace{-1.5em}

SAM is inherently a conditioned segmentator, so no architecture adaptations are needed.
The key challenge is prompting -- how to select keypoint prompts ($f(\mathcal{K}_i)$) and whether to prompt with bounding box ($g(\mathcal{B}_i)$).

SAM was trained with a maximum of 8 point prompts and fails with more, such as all 17 keypoints from COCO pose (\cref{fig:SAM-ablation-keypoints}).
The challenge is twofold: determining the number of keypoints and selecting them.
This chapter outlines our prompting method for a successful BBox-Mask-Pose loop and analyzes hyper-parameter effects on the loop.
Extensive ablation study on segmentation conditioned by pose with SAM is in the supplementary material.

\textbf{Visibility}.
Ideally, SAM should be prompted only by visible keypoints.
However, pose models estimate both visible and occluded keypoints and do not distinguish between them (with some exceptions, such as \cite{RethinkingVisibility}).
SAM can handle occluded keypoints if they are on the instance border but struggles if they are within another instance.
We approximate visibility by confidence and prompt only with keypoints above a confidence threshold; we have not been able to train, following \cite{RethinkingVisibility}, a reliable visibility predictor.

\textbf{Spread}.
To segment disconnected parts of an instance (for example, the legs of the background player in \cref{fig:BMP-visualization}), we maximize keypoints spread.
Selecting keypoints along the bounding box border provides a good spread, but SAM still needs at least one unambiguous keypoint to specify the instance.
We mimic human annotation by first choosing the most confident keypoint (analogous to a human’s initial click in the center) and then selecting keypoints to maximize spread.
To avoid redundancy, we select at most one facial keypoint (an eye or the nose).

\textbf{Bounding box}.
Another question is whether to use bounding boxes (\cref{fig:SAM-ablation-bbox}).
Experiments with ground truth boxes show that bounding boxes improve mask quality, but the situation changes with detected bounding boxes, especially in multi-body scenarios.
The detector may only capture part of an instance or merge two instances.
Prompting SAM with detected boxes restricts it to the detected area, limiting its ability to correct detection errors.
Conversely, SAM without a bounding box can “explore” undetected areas but loses precision within the bounding box.
Since detection correction is critical for BMP success, we prompt SAM without a detected bounding box.
Prompting with bounding box would be useful for final mask refinement after the BMP loop when bounding boxes are stable.

The keypoint selection algorithm is summarized in \cref{alg:keypoint-selection}.
It maximizes keypoint spread similar to KMeans++ initialization \cite{kmeans++}, factoring in keypoint confidence.
We used 6 positive keypoints for each instance ($N_{max}$) and confidence threshold $T_c = 0.5$.

Our experiments suggest that automatically selected keypoints have a different distribution from human-annotated prompts.
Human annotators intuitively understand the scene, and SAM generally performs better with human prompts than with automated keypoint selection.
Pose keypoints tend to lie on the borders and extremes of the instance, whereas humans often click in the middle of the instance.
By choosing visible, high-confidence, and spread keypoints, we partially simulate human prompting.
Although automated prompts do not match human effectiveness, the BBox-Mask-Pose loop still improves segmentation, and pose-prompted SAM outperforms bounding-box-prompted SAM.

\textbf{Pose-Mask consistency}.
When incorrect keypoints are selected during prompting, SAM’s segmentation mask may be worse than the original detector mask.
After mask generation, we measure the \textit{pose-mask consistency} of both the original detector mask and the mask refined by SAM.
Pose-mask consistency (\textit{P-Mc}) is defined as:
\vspace{-2em}%
\begin{equation}
    \text{P-Mc} = \frac{|k_p^+|}{| k_p|} + \frac{| k_n^-|}{| k_n|}
    \label{eq:PMc}
    \vspace{-2em}
\end{equation}
where $k_p$ represents the positive keypoints of the instance, and $k_n$ represents negative keypoints (those from other instances in the image).
$|k_p^+|$ is the number of positive keypoints inside the mask, while $|k_p^-|$ is the number of negative keypoints outside the mask.
Thus, pose-mask consistency measures the proportion of keypoints (both positive and negative) that are consistent with the mask.
If the refined mask has a lower \textit{P-Mc} than the previous mask, we discard it.
BBox-Mask-Pose discard approximately 15\% of SAM-refined masks.

Prompting with ground truth data behaves differently than with noisy estimated data.
As mentioned, the ground truth bounding box consistently improves the predicted mask. Similarly, ground truth data includes annotated visibility, allowing us to use only visible keypoints.
We prompted SAM with ground truth bounding boxes and poses when generating pseudo ground truth for AIC and MPII to train MaskPose.
For an extensive ablation study on prompting with ground truth or detections, see the supplementary material.

\setlength{\textfloatsep}{0pt}
\begin{algorithm}[tb]
    \SetKwData{Left}{left}
    \SetKwData{This}{this}
    \SetKwData{Up}{up}
    \SetKwFunction{Union}{Union}
    \SetKwFunction{FindCompress}{FindCompress}
    \SetKwInOut{Input}{Inputs}\SetKwInOut{Output}{Output}
    \SetKwComment{comment}{\#}{}
    \Input{Set ot detected keypoints $K$,\\Confidence threshold $T_c$,\\Max number of keypoint $N_{max}$}
    \BlankLine
    \Output{Set of selected keypoints $K_s$}
    \BlankLine
    
    Select keypoints from $K$ with confidence $\geq T_c$ \\
    Sort keypoints in $K$ by confidence \\
    $K_s \gets \emptyset$ \\
    Select the most confident keypoint into $K_s$ \\
    
    \While {$\text{len}(K_s) < N_{max}$}{
        $k \gets$ keypoint from $K$ furthest to $K_s$ \\
        Add $k$ to $K_s$
    }
    \textbf{return} $K_s$
    \caption{SAM prompts selection $f(\mathcal{K}_i)$}
    \label{alg:keypoint-selection}
\end{algorithm}
\setlength{\textfloatsep}{8pt}

\subsection{Closing the ``circle''}
\label{sec:method-loop}

With all three models adapted for mutual conditioning, we establish a closed iterative loop.
\begin{align}
    \big( \mathcal{B}_i, \mathcal{M}_i \big) &= \mathcal{D} \big( \mathcal{I} \! \odot \! (1 - \bigcup_{i} \mathcal{M}_i  ) \big)
    \label{eq:BMP_step1} \\
    \mathcal{K}_i &= \mathcal{P}_{\alpha} \big( \alpha \mathcal{I} + (1{-}\alpha)(\mathcal{I} \! \odot \! \mathcal{M}_i ), \mathcal{B}_i \big)
    \label{eq:BMP_step2} \\
    \big( \mathcal{B}_i, \mathcal{M}_i \big) &= \mathcal{S} \big( \mathcal{I}, f(\mathcal{K}_i), g(\mathcal{B}_i)  \big) 
    \label{eq:BMP_step3}
\end{align}

As shown in \cref{fig:BMP-visualization}, the detector conditions MaskPose (\cref{eq:BMP_step2}), which in turn conditions SAM2 segmentation (\cref{eq:BMP_step3}).
The loop completes by masking out processed instances and rerunning the detector (\cref{eq:BMP_step1}).

Each BBox-Mask-Pose iteration masks out more of the image, and when all instances are masked, the detector no longer identifies new instances, ending the loop.
In practice, the user can manually set the number of iterations, as later iterations yield diminishing performance gains.

To minimize duplicate detections, we use two forms of non-maximum suppression (NMS): bounding box NMS in the detector and pose NMS in the pose estimator.
We apply both with standard settings. Bounding box NMS with intersection-over-union (IoU) at 0.3 and pose NMS with object-keypoint-similarity (OKS) at 0.9.
If valid detections are mistakenly suppressed, they are re-detected in the next BMP loop iteration. 



\section{Results}
\label{sec:results}

\subsection{Implementation details}
\label{sec:results-implementation}

RTMDet-L \cite{RTMDet} is used in the BMP loop.
We fine-tuned RTMDet with instance-removal augmentation for 10 epochs on COCO-human, to enable it to ignore already-processed instances.
The same detector was used in top-down model experiments for a fair comparison.

MaskPose builds on ViTPose \cite{ViTPose} with the same training setup: 210 epochs on COCO, AIC and MPII. 
Since MPII and AIC lack ground truth segmentation, we generate pseudo ground truth using SAM2, prompted with ground truth bounding boxes and visible keypoints.

The Segment Anything Model (SAM) is used as is in version \textit{sam2-hiera-base+} with post-processing settings: max\_hole\_area at 10 and max\_sprinkle\_area at 50.
Each instance is processed independently. 

\subsection{Comparison with SOTA}
\label{sec:results-SOTA}

\begin{table}[tb]
    \centering
    \begin{tabular}{@{} l|| c | c }
        \toprule
        \multirow{2}{*}{Model} & OCHuman       & COCO    \\
                                & test AP       & val AP  \\

        \midrule
        DEKR       \cite{DEKR}  &  36.5           & 71.0   \\
        HQNet R-50$^{\mathparagraph}$ \cite{HQNet} &  40.0           & 69.5 \\
        CID-w48    \cite{CID} &  45.0           & 68.9 \\
        
        BUCTD \cite{BUCTD} &  47.4           & 74.8    \\
        
        \midrule
        Sapiens 0.3b \cite{Sapiens} &  41.3           & 66.1    \\
        MIPNet$^{\dagger}$   \cite{MIPNet} &  42.5           & 76.3  \\
        ViTPose-B \cite{ViTPose} &  42.6           & \underline{76.4}  \\ 

        \textbf{MaskPose-b}               &  45.0           & \textbf{76.5}  \\ 
        
        \midrule
        BUCTD $2\times$ \cite{BUCTD}  &  \underline{48.3}           & ---$^{\ddagger}$   \\
        

        \textbf{BBox-Mask-Pose} $1\times$                &  46.6           & \textbf{76.5}  \\


        \textbf{BBox-Mask-Pose} $2\times$              &  \textbf{49.2}           & \textbf{76.5}   \\        
        \bottomrule
    \end{tabular}
    \caption{
    \textbf{Pose estimation -- comparison with state-of-the-art}.
    Best results in bold, second best underlined.
    Results of detection-free (top), top-down  (middle) and iterative (bottom) methods.
    Top-down methods used detections from RTMDet-L \cite{RTMDet},  except MIPNet$^\dagger$ which reports results from \cite{MIPNet}.
    $\ddagger$ BUCTD $2\times$ result on COCO not reported \cite{BUCTD}.
    \vspace{0.2\baselineskip}
    $\mathparagraph$\cite{HQNet} ignores \textit{small} instances in~COCO.
    Summary: MaskPose improves ViTPose and it sets the new SOTA for top-down approaches.
    BMP further improves on MaskPose and set the SOTA for OCHuman while keeping SOTA on~COCO.  
    }
    \label{tab:SOTA-comparison-pose}
\end{table}

\begin{table}[tb]
    \centering
    \begin{tabular}{@{} l||c c }
        \toprule
        \multirow{2}{*}{Model} & \multicolumn{2}{c}{OCHuman test}  \\
                               &     bbox AP  &  mask AP            \\
        \midrule

        
        HRNet \cite{HRNet}  &   27.1       &  19.4               \\
        ConvNeXt  \cite{ConvNeXt} &   29.4       &  20.4               \\
        
        HQNet R-50%
        \tablefootnote{\cite{HQNet} also reports version with ViT-L backbone with better results. Its results could not be replicated as the authors do not provide weights.}
        \cite{HQNet} &   29.5       &  31.1              \\

        CoDETR SWIN-L$^{\ddagger}$ \cite{coDETR} &   29.6       &  ---               \\
        RTMDet-L \cite{RTMDet} &   30.0       &  26.5               \\

        \midrule
        Occlusion C\&P$^{\ddagger}$ \cite{HumanPaste} &   ---      &  28.3               \\
        ExPoSeg$^{\ddagger}$ \cite{PoSeg} &   ---      &  26.8               \\
        Crowd-SAM$^{\ddagger}$ \cite{CrowdSAM} &   ---       &  \underline{31.4}               \\

        \midrule
        \textbf{BBox-Mask-Pose} $1\times$             &   \underline{32.4}       &  30.2               \\

        \textbf{BBox-Mask-Pose} $2\times$             &   \textbf{35.9}       &  \textbf{34.0}               \\
        \bottomrule
    \end{tabular}
    \caption{
    \textbf{Detection and instance segmentation -- comparison with state-of-the-art}.
    Best results in bold, second best underlined.
    Results of COCO-trained detectors (top), segmentation models relyiong on previous detections or poses (middle).
    Models with~$\ddagger$ estimate either masks or report detection AP.
    Note that even CoDETR, a huge COCO SOTA model, struggles with multi-body scenes.
    BMP$\,2\times$ improves detection of RTMDet \cite{RTMDet} setting a new SOTA on OCHuman dataset.
    Qualitative results are in \cref{fig:examples-solved}.
    }
    \label{tab:SOTA-comparison-det}
    \vspace{-1em}
\end{table}

\begin{table*}[tb]
    \centering
    \begin{tabular}{@{}l||l|l|l|l|l||l}
        \toprule
        bbox AP @ max\_IoU    & 0.0 -- 0.2 & 0.2 -- 0.4 & 0.4 -- 0.6 & 0.6 -- 0.8 & 0.8 -- 1.0 & mAP \\
        \midrule
        RTMDet-L  & 16.9       & 0.1       & 20.4       & 15.7       & 8.7        & 31.1 \\
        BBox-Mask-Pose $2\times$    & 18.1 \impr{(+1.2)}  & 0.2 \impr{(+0.1)}  & 21.4  \impr{(+1.0)}  & 21.5  \impr{(+5.8)} & 10.7  \impr{(+2.0)}  & 35.7 \impr{(+4.6)} \\
        

        
        \bottomrule
    \end{tabular}
    \caption{
    \textbf{BMP's effectiveness for people with high overlap} on OCHuman-val.
    BMP improves detection especially in multi-body scenarios with big bbox overlap.
    Traditional detectors like RTMDet often merge two individuals into one instance or ignore the background individual.
    BMP resolves the issues with instance understanding through pose estimation.
    See e.g. the detection errors~in~\cref{fig:examples-solved}.
    }
    \label{tab:det-ablation-study}
\end{table*}


\begin{table}[tb]
    \centering
    \begin{tabular}{@{}c c c c | c |r r}
        \toprule
        det & pose   & SAM    & pose   & loops      & bbox  & pose  \\
        \midrule
        \cmark & \cmark & \cmark & \xmark & $1\times$  & 31.1     & 45.3     \\
        \cmark & \cmark & \cmark & \xmark & $2\times$  & \textbf{32.1}     & \textbf{48.6}     \\
        \midrule
        \cmark & \cmark & \cmark & \textcolor{magenta}{\cmark} & $1\times$  & 31.1     & 46.4    \\
        \cmark & \textcolor{magenta}{\xmark} & \cmark & \xmark & $2\times$  & \underline{31.9}     & \underline{47.3}     \\
        \cmark & \cmark & \textcolor{magenta}{\xmark} & \xmark & $2\times$  & 30.8     & 47.0    \\
        \bottomrule
    \end{tabular}
    \caption{
    \textbf{Ablation study} of BBox-Mask-Pose components evaluated on OCHuman-val.
    Bbox and pose evaluated with AP.
    The sum of trainable parameters approximates computational complexity.
    First row corresponds to BMP $1\times$, second to BMP $2\times$.
    }
    \label{tab:ablation-study}
\end{table}

\textbf{Pose estimation}.
\cref{tab:SOTA-comparison-pose} compares pose estimation performance on the OCHuman and COCO datasets.
MaskPose improves the ViTPose \cite{ViTPose} baseline from 42.6 to 45.0 AP by mask conditioning, making it a new SOTA among top-down methods.
\edit{
BMP $1\times$ yields better results than MaskPose on OCHuman (45.0 $\rightarrow$ 46.6), because it uses fine-tuned detector from \cref{sec:Method-detection}.
Otherwise BMP $1\times$ outputs the same pose as MaskPose.
}
BMP $2\times$ further increases BMP $1\times$ performance from 46.6 to 49.3 AP through iterative conditioning between masks and poses.
BMP sets the new SOTA performance on OCHuman, beating BUTCD \cite{BUCTD}.
BMP and MaskPose perform similarly on COCO, as the detector captures nearly all instances in the first pass, with only a few additional detections in the second iteration.

Additionally, the numbers could improve with bigger models (ViTPose-h, RTMDet-x, SAM2.1-large) and additional bells and whistles (e.g. BUCTD).
BUCTD could either refine MaskPose's keypoints or replace MaskPose within the BMP loop as it conditions pose estimation on bottom-up poses while MaskPose is conitioned on masks.

Experiments show that performance plateaus after two iterations, similar to BUCTD.
Further iterations add computational cost without notable performance gains.

\textbf{Detection and segmentation}.
\cref{tab:SOTA-comparison-det} shows BMP detection and segmentation performance on the OCHuman dataset.
BMP $1\times$ improves the RTMDet pipeline by refining bounding boxes and segmentation masks using pose-prompted SAM, as illustrated in \cref{fig:BMP-visualization}.
BMP $2\times$ further improves detection and segmentation through re-detection of background instances in images with masked-out instances, as shown in \cref{fig:examples-solved}.
BBox-Mask-Pose sets a new SOTA on OCHuman detection and segmentation beating both object detectors and pose-conditioned segmentators such as ExPoSeg \cite{PoSeg}.
\edit{Experiments on CIHP \cite{CIHP} are in the Suppl.}

\textbf{Detection accuracy in multi-body scenarios}.
\cref{tab:det-ablation-study} shows that the detection performance is improved most in scenarios with a high bbox overlap.
For each GT instance, we calculate its highest IoU with other GT instances (max\_IoU) and split the OCHuman dataset accordingly.
Detections cannot be split accordingly as high inter-detection overlap could be both multi-body scenarios and false positives.
Therefore, AP numbers are generally lower than for standard mAP metric but the comparison between models is fair.
Qualitative examples of improvement are in \cref{fig:examples-solved}.

\subsection{Ablation study}
\label{sec:results-ablation}

\textbf{Looping SAM and pose estimation}.
The third row of \cref{tab:ablation-study} shows a slight improvement in pose estimation when re-running pose on SAM-refined masks.
This pipeline, detect-pose-SAM-pose, is comparable to one BMP iteration as it cannot re-detect previously missed instances.
Formally, the experiment is chaining \cref{eq:BMP_step2,eq:BMP_step3} without \cref{eq:BMP_step1}.
SAM mask refinement improves MaskPose keypoint predictions, suggesting that an SAM-pose-SAM loop could further enhance the results.
However, the additional computational cost outweighs the gains, so we exclude it to keep BMP efficient.

\textbf{Prompting SAM only with bounding box}.
This approach effectively omits the pose estimation model (\cref{eq:BMP_step2}) from the loop, as SAM is prompted solely by the bounding box detected in the first step.
SAM refines the segmentation mask and updates the bounding box accordingly.
\cref{tab:ablation-study} shows that SAM alone improves performance over omitting SAM entirely (second-last and last rows).
Adding keypoints as prompts further boosts detection from 31.9 to 32.1 AP and pose estimation from 47.3 to 48.6 AP.

\textbf{Omitting SAM}.
When SAM (\cref{eq:BMP_step3}) is omitted from BMP, segmentation masks are provided only by the detector from \cref{eq:BMP_step1}.
This causes the detector to loop with itself without conditioning from masks or poses, often resulting in un-segmented body parts, such as missed limbs.
For example, in \cref{fig:BMP-visualization}, un-segmented legs of a background player could be detected as separate instances, as shown in \cref{fig:BMP-failures}.
In practice, omitting SAM resembles running a detector with a low non-maxima suppression (NMS) threshold, resulting in many false-positive bounding boxes.
This hinders detection performance, but slightly boosts pose accuracy.
Low-confidence poses minimally impact the COCO evaluation, as they do not deform the precision-recall curve in the AP computation.
That explain why looping the detector with itself still improves the pose.
However, using SAM improves detection from 30.8 to 32.1 AP and pose estimation from 47.0 to 48.6 AP, as shown in \cref{tab:ablation-study}.


\begin{table}[t]
    \centering
    \begin{tabular}{@{}l|r r r@{}}
        \toprule
        model & s/img &  params & pose mAP  \\
        \midrule
        RTMDet-L           & 0.03  &  57 M & ---  \\ 
        Sapiens 0.3b       & 1.95  & 336 M & ---  \\ 
        MaskPose-b         & 0.06  &  87 M & ---  \\ 
        SAM2-hiera-base+   & 0.47  &  81 M & ---  \\
        \midrule
        RTMDet-L + Sapiens 0.3b        & 2.03  & 393 M & 41.3 \\
        BBox-Mask-Pose $1\times$       & \textbf{0.56}  & 225 M & \underline{46.6} \\ 
        BBox-Mask-Pose $2\times$       & \underline{1.12}  & 369 M & \textbf{49.2} \\
        \bottomrule
    \end{tabular}
    \caption{
    \textbf{Runtime analysis} on OCHuman; s/img -- seconds per image.
    Measured on A-100 GPU with 40 GB.
    BMP~$2\times$ is almost two times faster than Sapiens while having better performance. 
    \vspace{-0.5em}
    }
    \label{tab:runtime-study}
\end{table}

\textbf{Computational complexity estimation}.
\cref{tab:runtime-study} compares BMP runtime to Sapiens 0.3b \cite{Sapiens}, a recent foundational model with 336M parameters.
Combined with RTMDet-L, it totals 393M parameters, surpassing the 369M of two BMP iterations.
Similarly, runtime analysis shows that BMP $2\times$ runs half the time while outperforming Sapiens on both COCO and OCHuman datasets.
The runtime analysis proves that multiple small specialized models are faster and achieve better performance than huge foundational models.
For complexity analysis of various components of the BMP loop, see the supplementary material. 

\section{Conclusions}
\label{sec:conclusions}

We present BBox-Mask-Pose (BMP), a method for detection, segmentation, and pose estimation in multi-body scenarios.
Part of the BMP loop, a new top-down model MaskPose, conditions pose estimation on predicted instance masks unlike prior approaches.
BMP integrates detector, MaskPose and (SAM) into a self-improving loop.
By conditioning each model on outputs from the others, BMP simultaneously improves detection, segmentation, and pose estimation and set a new SOTA on the OCHuman dataset in all three tasks.
Key findings are:

\begin{enumerate}

\item Conditioning the top-down pose model with masks and bounding boxes improves performance, especially in crowded scenes.

\item BMP demonstrates that explicit mutual conditioning between the detector, segmentator, and pose estimation models improve performance in all tasks.
Small specialized models give better results than large foundational models with shared features.
However, adapting these models for mutual conditioning is non-trivial.

\item BMP’s effectiveness diminishes after two iterations, with additional iterations offering little performance gain while increasing computational cost.

\item BMP sets the new SOTA on OCHuman while also matching the SOTA performance of top-down models on COCO.

\item Surprisingly, the Segment Anything Model proved the least effective component in BMP. Even though BMP segmentation is the new SOTA, automated SAM prompting falls short compared to human interaction and most of the errors come from incorrect masks.

\item The modular structure of BMP enables further performance gains by integrating improved models or adding BUCTD \cite{BUCTD} to the loop.

\end{enumerate}

\begin{figure}[tb]
    \centering
    \begin{subfigure}{0.31\linewidth}
        \centering
        \includegraphics[width=\textwidth,page=4]{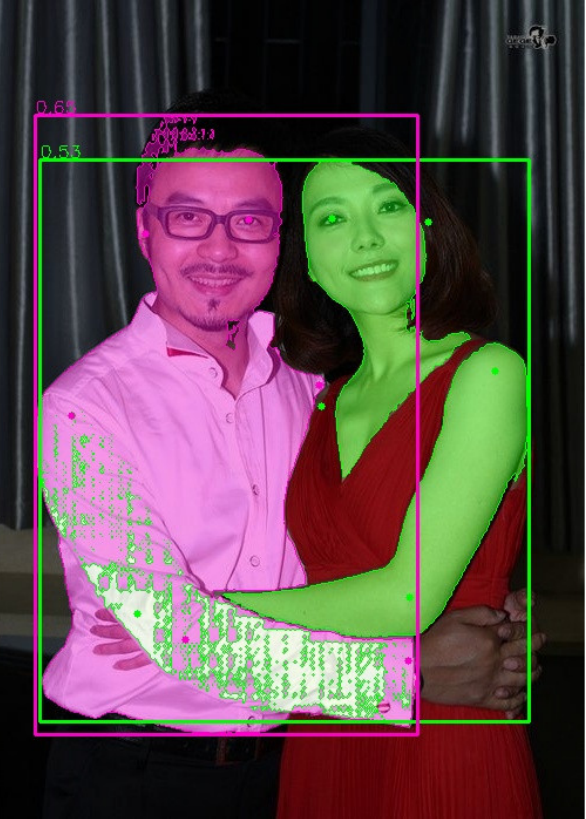}
        \caption{
        Segmenting only skin (green) 
        }
    \end{subfigure}
    \hfill
    \begin{subfigure}{0.31\linewidth}
        \centering
        \includegraphics[width=\textwidth,page=5]{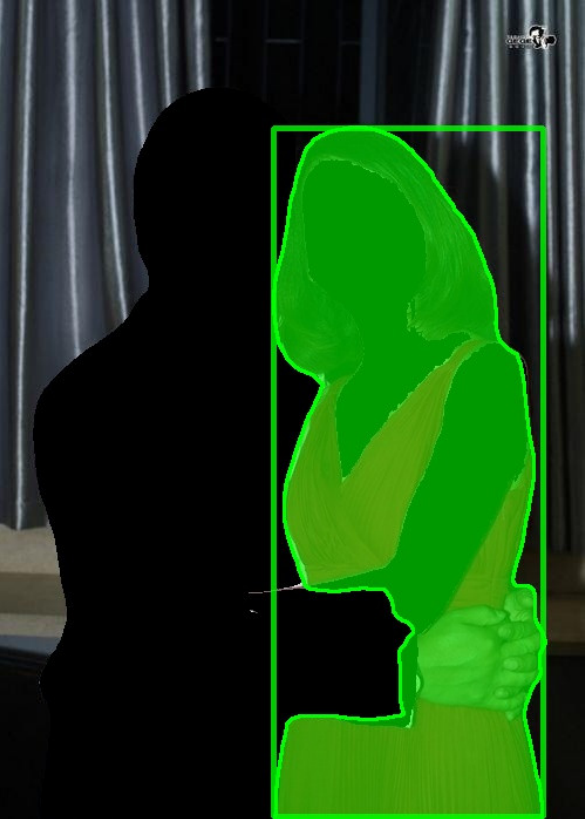}
        \caption{
        Re-detection of un-segmented clothes.
        }
    \end{subfigure}
    \hfill
    \begin{subfigure}{0.324\linewidth}
        \centering
        \includegraphics[width=\textwidth,page=12]{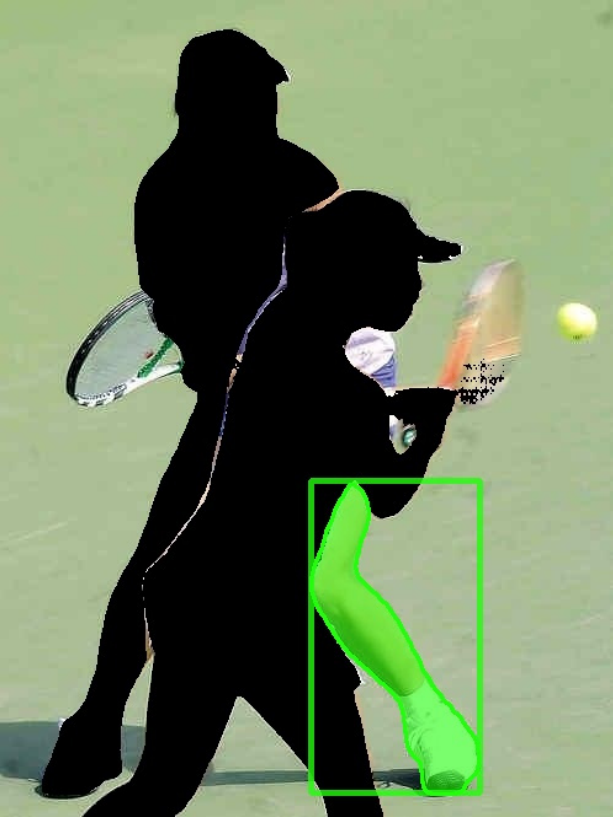}
        \caption{
        Missed limb is re-detected
        }
        \label{fig:BMP-failures-missed-limb}
    \end{subfigure}
    \caption{
    \textbf{Typical errors} of the BMP loop.
    The weakest part is SAM and its prompting with correct keypoints.
    \vspace{-0.5em}
    }
    \label{fig:BMP-failures}
\end{figure}

\textbf{Limitations} of BMP primarily involve imperfect SAM mask refinement.
When SAM is prompted with inaccurate keypoints (e.g., occluded or mislocalized), it has limited recovery ability, which can lead to masking out the wrong instances, preventing the detector from retrieving them.
We experimented with semi-transparent masking, as used in MaskPose, but found it ineffective.

A second limitation occurs when detecting in masked-out images.
If a foreground instance divides a background instance into disconnected parts, the detector often fails to connect these, generating multiple small bounding boxes for each segment.
Although pose NMS suppresses redundant detections, disconnected body parts remain separate.
Attempts to use data augmentation to improve detector robustness in such cases were unsuccessful.
Examples of these errors are included in \cref{fig:BMP-failures}.
More detailed analysis of SAM errors is provided in the supplementary material.

\textbf{Future work}.
Improving the efficiency of interactions between bounding boxes, masks, and poses is an area for exploration.
Foundational models aim to unify body representations at a feature level but lack the explicit constraints offered by different representations.
Although foundational models are non-iterative, their large size often results in longer inference times compared to smaller, specialized models.
Our findings indicate that explicit constraints within specialized models could improve performance while keeping the models smaller and faster.

%% file: sec/acknowledgements.tex
\textbf{Acknowledgements}.
This work was supported by
the Ministry of the Interior of the Czech Republic project No. VJ02010041, 
the Technology Agency of the Czech Republic project CEDMO 2.0 No. FW10010387, 
the European Union’s Digital Europe Programme under Contract No. 101158609, 
and the Czech Technical University student grant SGS23/173/OHK3/3T/13.

%% file: sec/supplementary.tex
\appendix
\maketitlesupplementary

\section{Prompting SAM ablation study}
\label{sec:prompting-sam}

\subsection{Setup}
\label{sec:prompting-sam-setup}

Here, we describe the ablation study on prompting SAM.
The study evaluates three metrics: detection improvement (bounding box; bbox), segmentation improvement (segm), and pose improvement (pose).
For all experiments, we use bounding boxes and segmentation masks from RTMDet-l and pose estimates from MaskPose as the baseline pipeline.
The experimental pipeline remains consistent throughout.

Detection and segmentation changes are evaluated on bounding boxes and segmentation masks refined by SAM, following the det-pose-SAM pipeline.
Pose estimation is assessed by re-running MaskPose on refined masks, forming a det-pose-SAM-pose pipeline, similar to the setup in \cref{tab:ablation-study}.

All experiments use \textit{RTMDet-l} \cite{RTMDet} as the detector, \textit{MaskPose-b} as the pose estimator, and \textit{sam2-hiera-base+} as the SAM2 \cite{SAM2} model.
Each experiment is assigned a specific name, listed in the leftmost column of the tables, for clear referencing. When experiments appear in multiple tables for comparison, their names remain consistent for easier cross-referencing.
Each result is highlighted in green or red depending on whether it improves or hinders performance compared to the RTMDet+MaskPose baseline.

\begin{figure}[tb]
    \centering
    \begin{subfigure}{0.49\linewidth}
        \centering
        \includegraphics[width=\textwidth,page=31]{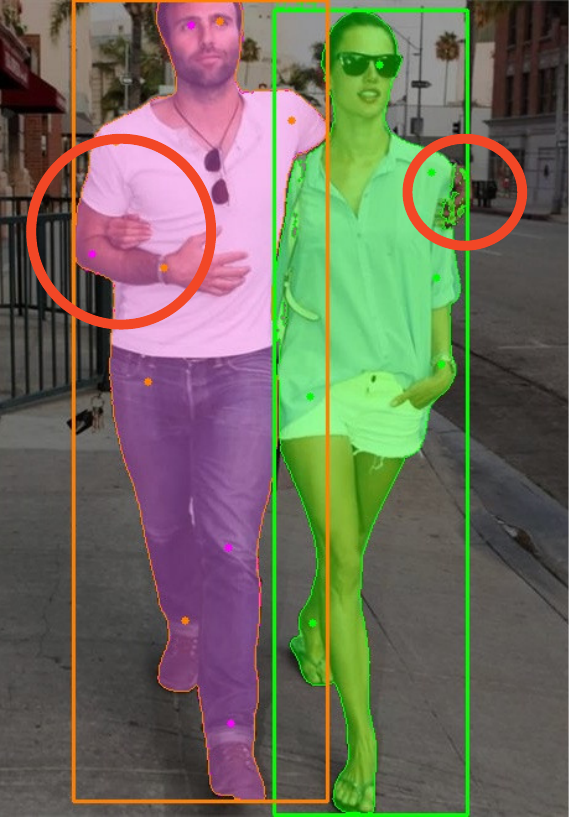}
        \caption{
        }
    \end{subfigure}
    \hfill
    \begin{subfigure}{0.49\linewidth}
        \centering
        \includegraphics[width=\textwidth,page=32]{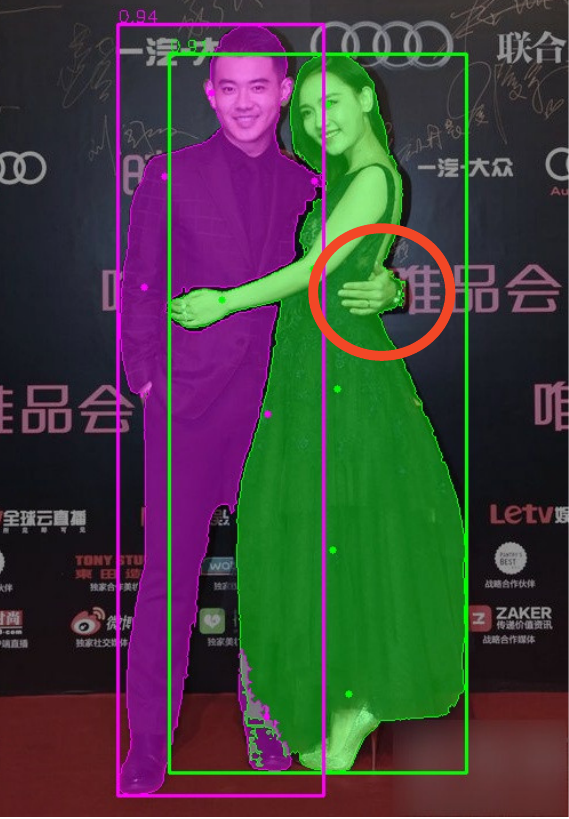}
        \caption{
        }
    \end{subfigure}
    \caption{
    Segmentation error involving a small number of pixels, like the
    circled hands, may have a large impact on detection accuracy measured by bounding box IoU.
    A detector returning  correct bounding boxes, which would be nearly identical for both persons especially in (a), can make segmentation of the two people very challenging.
    Improving detection may thus lead to decrease in  segmentation
    performance.
    Keypoints used for SAM prompting are marked (best viewed in zoom).
    }
    \label{fig:SUPPL-det-vs-segm}
\end{figure}

\textbf{Detection vs. segmentation}.
Before analyzing the results of the ablation study, we address a counterintuitive observation.
When refining masks on OCHuman, segmentation and detection often conflict; improvement in one can lead to a decrease in the other.
This is due to the focus on people with high overlap in the OCHuman dataset.
Many examples consist of a large area representing the main body and smaller, disconnected body parts.
Examples are shown in \cref{fig:SUPPL-det-vs-segm}.

When mask refinement focuses heavily on the main segment, segmentation scores improve, as missing disconnected parts has little impact on mask IoU.
Conversely, overly general prompting can cause SAM to merge both instances into one mask, creating a bounding box that may be more accurate than the original. Large masks merge instances, while small masks often miss disconnected body parts.

We prioritize detection, even though the goal is to improve all three metrics.
The mask refinement step in BBox-Mask-Pose must ensure that segmented masks adequately remove limbs during the mask-out step, as shown in \cref{fig:BMP-failures-missed-limb,fig:SUPPL-missed-limb}.
However, excessively large masks prevent decoupling of merged instances, as seen in \cref{fig:examples-solved-2g1d_merge}.
Thus, our aim is to improve detection without significantly hindering segmentation performance.

\subsection{Results}
\label{sec:prompting-sam-results}

\begin{table}[tb]
    \centering
    \begin{tabular}{@{}l | c c c c | r r r}
        \toprule
       name & batch  & bbox   & $\cplus$      & $\cminus$      & bbox       & segm       & pose \\  
        \midrule
        \multicolumn{5}{c|}{RTMDet \cite{RTMDet} + MaskPose}  & 31.1       & 27.1       & 45.3 \\
        \midrule
        
        A1  & \xmark & \cmark & 0      & 0      & \red{27.5} & \grnb{31.6} & \red{44.2} \\   
        A2  & \xmark & \cmark & 2      & 0      & \red{28.5} & \grnb{31.6} & \redb{44.3} \\  
        \rowcolor{LightBlue}
        A3  & \xmark & \cmark & 4      & 0      & \red{29.3} & \grn{30.9} & \red{44.0} \\    
        A4  & \xmark & \cmark & 6      & 0      & \red{30.4} & \grn{29.0} & \red{43.6} \\    
        A5  & \xmark & \cmark & 8      & 0      & \grnb{31.4} & \red{26.9} & \red{43.5} \\   
        \midrule
        B1  & \xmark & \xmark & 1      & 0      & \red{ 2.5} & \red{ 2.8} & \red{12.6} \\    
        B2  & \xmark & \xmark & 2      & 0      & \red{20.5} & \red{20.6} & \red{39.8} \\    
        B3  & \xmark & \xmark & 4      & 0      & \grn{31.6} & \grnb{29.1} & \redb{43.5} \\  
        \rowcolor{LightBlue}
        B4  & \xmark & \xmark & 6      & 0      & \grn{32.2} & \grn{27.3} & \red{42.7} \\    
        B5  & \xmark & \xmark & 8      & 0      & \grnb{32.5} & \red{26.0} & \red{42.1} \\   
        B6  & \xmark & \xmark & 10     & 0      & \grn{32.2} & \red{24.2} & \red{41.4} \\    
        \bottomrule
    \end{tabular}
    \caption{
    Ablation study on prompting SAM \cite{SAM2} with varying positive keypoints ($\cplus$) on OCHuman-val.
    Best results for each metric highlighted in \textbf{bold}; best method for BMP \colorbox{LightBlue}{highlighted in blue}.
    \grn{Green text} indicates improvement over the baseline, \red{red text} indicates a decline.
    Detection and segmentation often conflict (\cref{fig:SUPPL-det-vs-segm}).
    More keypoints improve segmentation (including incorrect masks) and bounding box detection, but increase segmentation errors.
    Pose remains stable but suffers from both wrong segmentation (guidance errors) and wrong detection (crop errors).
    }
    \label{tab:sam-ablation-study-num-pos-kpts}
\end{table}

\textbf{Bounding box}.
The question of whether to prompt SAM with a bounding box is addressed in \cref{tab:sam-ablation-study-num-pos-kpts}, with examples provided in \cref{fig:SAM-ablation-bbox}.
When the bounding box is accurate, or nearly so, it significantly improves segmentation quality.
However, when the bounding box is incorrect, such as missing parts of an occluded person (\cref{fig:BMP-failures-missed-limb}), prompting restricts mask refinement to the given bounding box, reducing the chance of recovery.

In the final version of BBox-MaskPose, we do not use bounding box prompting, as we prioritize SAM’s ability to explore and detect previously missed body parts (\cref{fig:SUPPL-good-fix}).
However, when bounding boxes are reliable, prompting with them can further refine segmentation and pose estimation, yielding improved results, as shown in \cref{tab:ablation-study} in \cref{sec:results-ablation}.
Bounding box prompting is also advantageous when ground truth bounding boxes are available.

\begin{table}[tb]
    \centering
    \begin{tabular}{@{}l | c c c c | r r r}
        \toprule
       name & batch  & bbox   & $\cplus$  & $\cminus$    & bbox       & segm       & pose \\  
        \midrule
        \multicolumn{5}{c|}{RTMDet \cite{RTMDet} + MaskPose}  & 31.1       & 27.1       & 45.3 \\
        \midrule
        \rowcolor{LightBlue}
        A3   & \xmark & \cmark & 4      & 0     & \red{29.3} & \grnb{30.9} & \red{44.0} \\    
        C1  & \xmark & \cmark & 4      & 1     & \red{29.5} & \grn{30.5} & \red{44.3} \\     
        C2  & \xmark & \cmark & 4      & 3     & \redb{29.8} & \grn{28.2} & \redb{44.2} \\   
        C3  & \cmark & \cmark & 4      & --    & \red{29.3} & \grnb{30.9} & \red{44.0} \\    
        \midrule
        \rowcolor{LightBlue}
        B4   & \xmark & \xmark & 6      & 0     & \grnb{32.2} & \grnb{27.3} & \red{42.7} \\   
        C4  & \xmark & \xmark & 6      & 1     & \red{29.9} & \red{23.8} & \red{43.6} \\     
        C5  & \xmark & \xmark & 6      & 3     & \red{27.5} & \red{19.2} & \redb{44.1} \\    
        C6  & \cmark & \xmark & 6      & --    & \grnb{32.2} & \grnb{27.3} & \red{42.7} \\   
        \bottomrule
    \end{tabular}
    \caption{
    Ablation study on prompting SAM \cite{SAM2} with varying negative keypoints ($\cminus$) on OCHuman-val.
    Best results for each metric in \textbf{bold}; best method for BMP \colorbox{LightBlue}{highlighted in blue}.
    \grn{Green text} indicates improvement over the baseline, \red{red text} indicates a decline.
    Adding negative keypoints to bounding boxes hinders segmentation but slightly improves detection.
    Without bounding boxes, negative keypoints degrade both detection and segmentation.
    Processing all image instances simultaneously (batch) gives the same or worse results.
    }
    \label{tab:sam-ablation-study-num-neg-kpts}
\end{table}

\textbf{Number of positive keypoints} ($\cplus$).
\cref{tab:sam-ablation-study-num-pos-kpts} evaluates the effect of using different numbers of keypoints for prompting. 

In the top section, which includes bounding box prompts, using more keypoints increases the likelihood of confusing the model, leading to a drop in segmentation quality.
However, more keypoints also increase the chance of expanding the mask beyond the bounding box, which improves detection.
In particular, using 8 keypoints as positive prompts slightly outperforms the original baseline in detection.

The second section, without bounding box prompts, highlights that too few keypoints fail to define the instance adequately, causing both detection and segmentation to fail catastrophically.
The best segmentation results occur with 4 keypoints, while detection performs best with 8. We chose 6 keypoints as a middle ground, balancing strong detection performance with slightly improved segmentation.

\textbf{Number of negative keypoints} ($\cminus$).
SAM2 provides two methods for negative prompting: explicit negative prompts and batch processing of all instances in the image.
For explicit negative prompts, we identify the closest keypoint from other instances in the same image, provided it has confidence above a specified threshold.

\cref{tab:sam-ablation-study-num-neg-kpts} evaluates the impact of negative keypoint prompts.
The top section examines adding negative prompts to 4 positive prompts and a bounding box.
Negative prompts slightly improve detection quality, but significantly reduce segmentation quality.
Given the trade-off, the decrease in segmentation outweighs the minor improvement in detection, so we avoid using negative keypoints in this setup.

The bottom section evaluates the effect of negative prompts without a bounding box prompting.
Here, adding negative keypoints decreases both detection and segmentation performance, making it ineffective for this configuration.

\textbf{Batch processing}.
\cref{tab:sam-ablation-study-num-neg-kpts} also evaluates the impact of batch processing, where SAM is prompted with multiple instances simultaneously.
In this approach, SAM outputs non-overlapping masks for each prompted instance, ensuring that no mask is a subset of another.
Although this behavior is logical, batch processing consistently produced the same or slightly lower results compared to single-instance processing in all our experiments.

We chose to stick with single-instance processing, as it likely allows the model to optimize better for one instance at a time, even if the resulting masks may overlap.
Overlaps could be resolved in a post-processing step using pose information.

\begin{table*}[tb]
    \centering
    \begin{tabular}{@{}l  c c c c c c c c c | r r r}
        \toprule
       name  & batch  & bbox   & $\cplus$      & $\cminus$      & $T_c$ & sel.    & ext. bbox & P-Mc   & bbox by IoU & bbox       & segm       & pose \\  
        \midrule
        \multicolumn{10}{c|}{RTMDet \cite{RTMDet} + MaskPose}                                                     & 31.1       & 27.1       & 45.3 \\
        \midrule
        
        \multicolumn{13}{l}{\hspace{-0.21cm}Confidence threshold $T_c$} \\
        D1  & \xmark & \xmark & 6      & 0      & 0.8  & c+d    & ---       & \xmark & \xmark      & \red{29.9} & \grn{27.2} & \red{42.1} \\   
        B4   & \xmark & \xmark & 6      & 0      & 0.5  & c+d    & ---       & \xmark & \xmark      & \grn{32.2} & \grn{27.3} & \red{42.7} \\   
        D2  & \xmark & \xmark & 6      & 0      & 0.4  & c+d    & ---       & \xmark & \xmark      & \grn{32.4} & \grn{27.6} & \red{43.1} \\   
        \rowcolor{LightBlue}
        D3  & \xmark & \xmark & 6      & 0      & 0.3  & c+d    & ---       & \xmark & \xmark      & \grnb{32.7} & \grn{27.9} & \red{43.3} \\  
        D4  & \xmark & \xmark & 6      & 0      & 0.2  & c+d    & ---       & \xmark & \xmark      & \grn{32.5} & \grnb{28.3} & \redb{43.6} \\ 
        D5  & \xmark & \xmark & 6      & 0      & 0.1  & c+d    & ---       & \xmark & \xmark      & \grn{32.5} & \grn{28.2} & \redb{43.6} \\  
        
        \vspace{-0.3cm} \\
        \multicolumn{13}{l}{\hspace{-0.21cm}Selection method} \\
        \rowcolor{LightBlue}
        D3  & \xmark & \xmark & 6      & 0      & 0.3  & c+d    & ---       & \xmark & \xmark      & \grnb{32.7} & \grnb{27.9} & \red{43.3} \\ 
        E1 & \xmark & \xmark & 6      & 0      & 0.3  & c      & ---       & \xmark & \xmark      & \red{29.7} & \red{26.2} & \redb{45.0} \\  
        E2 & \xmark & \xmark & 6      & 0      & 0.3  & d      & ---       & \xmark & \xmark      & \grn{34.6} & \red{20.6} & \red{36.8} \\   
        
        \vspace{-0.3cm} \\
        \multicolumn{13}{l}{\hspace{-0.21cm}Extended bounding box} \\
        \rowcolor{LightBlue}
        F1  & \xmark & \cmark & 4      & 0      & 0.3  & c+d    & \xmark    & \xmark & \xmark      & \red{29.3} & \grnb{31.1} & \red{44.1} \\  
        F2 & \xmark & \cmark & 4      & 0      & 0.3  & c+d    & \cmark    & \xmark & \xmark      & \red{29.7} & \grn{31.0} & \red{44.1} \\   
        
        \vspace{-0.3cm} \\
        \multicolumn{13}{l}{\hspace{-0.21cm}Pose-Mask consistency} \\
        \rowcolor{LightBlue}
        D3  & \xmark & \xmark & 6      & 0      & 0.3  & c+d    & ---       & \xmark & \xmark      & \grnb{32.7} & \grn{27.9} & \red{43.3} \\  
        G1 & \xmark & \xmark & 6      & 0      & 0.3  & c+d    & ---       & \cmark & \xmark      & \red{30.9} & \grnb{31.1} & \redb{45.0} \\ 

        \vspace{-0.3cm} \\
        \multicolumn{13}{l}{\hspace{-0.21cm}Bounding box by max\_IoU} \\
        \rowcolor{LightBlue}
        D3  & \xmark & \xmark & 6      & 0      & 0.3  & c+d    & ---       & \xmark & \xmark      & \grnb{32.7} & \grn{27.9} & \red{43.3} \\  
        F1  & \xmark & \cmark & 4      & 0      & 0.3  & c+d    & \xmark    & \xmark & \xmark      & \red{29.3} & \grnb{31.1} & \redb{44.1} \\  
H1  & \xmark & \xmark/\cmark & 6/4      & 0      & 0.3  & c+d    & \xmark    & \xmark & \cmark      & \red{29.7} & \grn{30.1} & \red{43.9} \\ 

        \vspace{-0.3cm} \\
        \multicolumn{13}{l}{\hspace{-0.21cm}Final methods} \\
        \rowcolor{PastelGreen}
        D3  & \xmark & \xmark & 6      & 0      & 0.3  & c+d    & ---       & \xmark & \xmark      & \grnb{32.7} & \grn{27.9} & \red{43.3} \\   
        \rowcolor{PastelGreen}
        J1   & \xmark & \xmark/\cmark & 6/4    & 0      & 0.5  & c+d    & \cmark    & \cmark & \cmark      & \red{29.2} & \grn{31.1} & \grn{46.3} \\  
        \bottomrule
    \end{tabular}
    \caption{
    Ablation study on prompting SAM \cite{SAM2} with varying confidence thresholds ($T_c$), keypoint selection methods (sel.), and additional techniques on OCHuman-val.
    Best results for each metric in \textbf{bold}; best method for BMP \colorbox{LightBlue}{highlighted in blue}.
    \grn{Green text} indicates improvement over the baseline, \red{red text} indicates a decline.
    Final methods used in BBox-Mask-Pose are \colorbox{PastelGreen}{highlighted in green}.
    Two different methods used: one for the BMP loop, another for mask and pose refinement.
    }
    \label{tab:sam-ablation-study-other}
\end{table*}

\textbf{Confidence threshold} ($T_c$).
The top part of \cref{tab:sam-ablation-study-other} examines the effect of varying the confidence threshold $T_c$ for selecting keypoints as prompts.
Lower thresholds select keypoints with greater variability but increase the risk of using incorrectly estimated keypoints.
The best results are achieved with a threshold of $T_c = 0.3$, which aligns with its common use in heatmap-based pose estimation models.

Interestingly, a lower threshold ($T_c = 0.1$) outperforms a higher threshold ($T_c = 0.8$), suggesting that variability is more important than strictly ensuring keypoint correctness.
This may indicate that SAM is either robust to incorrect prompts (which we find unlikely) or that confidence is not a reliable metric for evaluating keypoint accuracy.
As human pose estimation models are often overconfident, using self-estimated OKS from \cite{Calibration} could likely yield better results than relying on confidence.

\textbf{Selection method} (sel.).
We compare three methods for selecting keypoints as prompts.
The first method, confidence-only (c), sorts keypoints by confidence and selects the top N most confident ones.
The second, distance-only (d), selects the N keypoints farthest from the center of the bounding box.
The third method, described in \cref{sec:Method-segmentation}, combines confidence and distance (c+d).

The second part of \cref{tab:sam-ablation-study-other} shows that combining confidence and distance (c+d) outperforms either approach alone, providing superior results.

\textbf{Extending bounding box}.
Experiment F2 in \cref{tab:sam-ablation-study-other} explores the idea of extending the bounding box when using it for prompting.
If selected keypoints fall outside the bounding box, it is extended to include all prompt keypoints.
This ensures that no positive prompt lies outside the bounding box.

The results show that extending the bounding box slightly improves the detection accuracy while maintaining segmentation and pose estimation performance when using the bounding box.
This approach is not applicable when prompting without a bounding box.

\textbf{Pose-Mask consistency} (P-Mc).
Experiment G1 in \cref{tab:sam-ablation-study-other} evaluates the effect of Pose-Mask Consistency (P-Mc), as described in \cref{sec:Method-segmentation}.
P-Mc significantly improves segmentation and pose estimation, but reduces detection performance.
As a result, it is highly effective for refining masks and poses when the bounding box is approximately correct but not suitable for use in the iterative BBox-Mask-Pose loop.

\begin{figure}[tb]
    \centering
    \begin{subfigure}{0.49\linewidth}
        \centering
        \includegraphics[width=\textwidth,page=48]{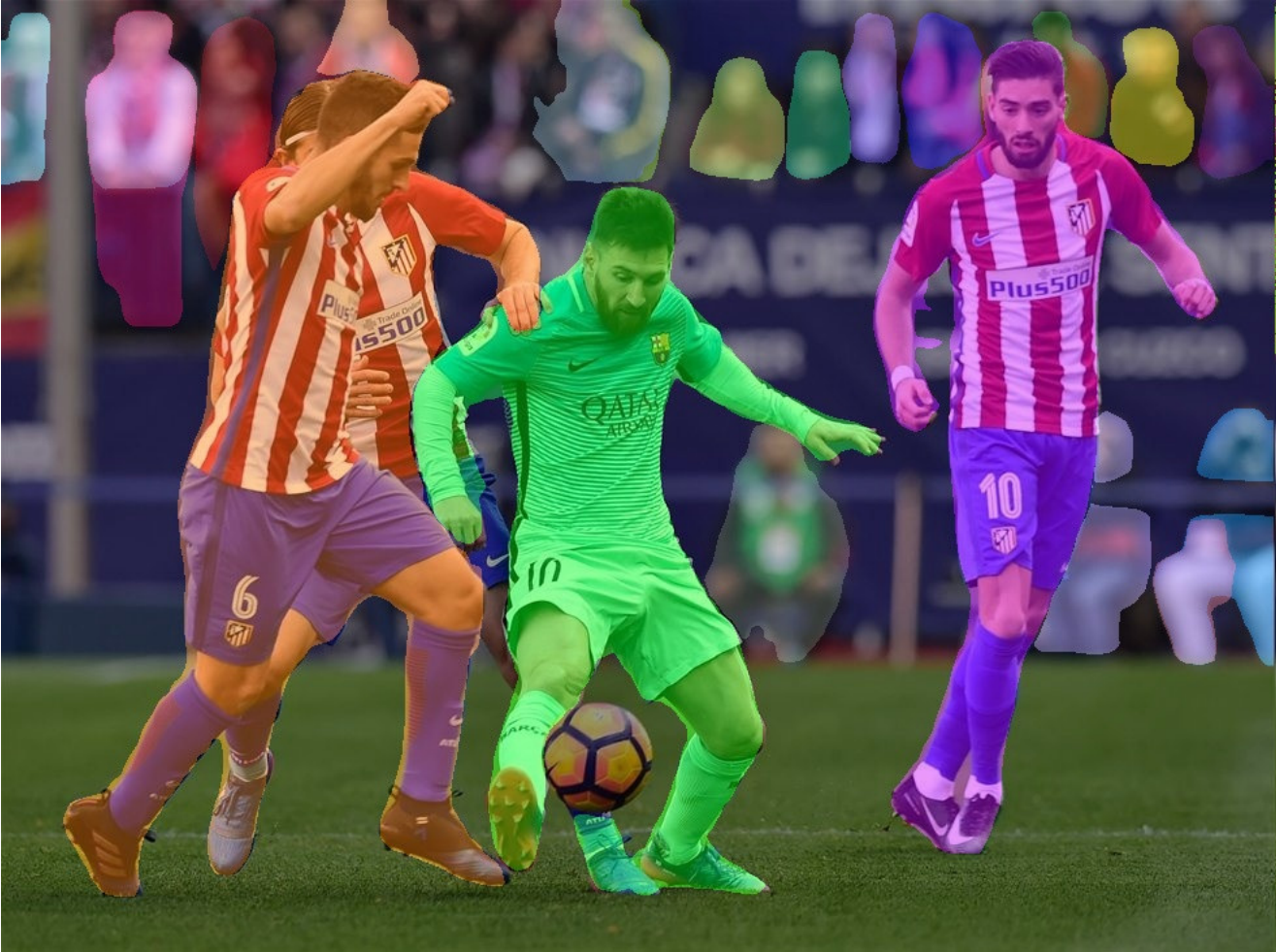}
    \end{subfigure}
    \hfill
    \begin{subfigure}{0.49\linewidth}
        \centering
        \includegraphics[width=\textwidth,page=49]{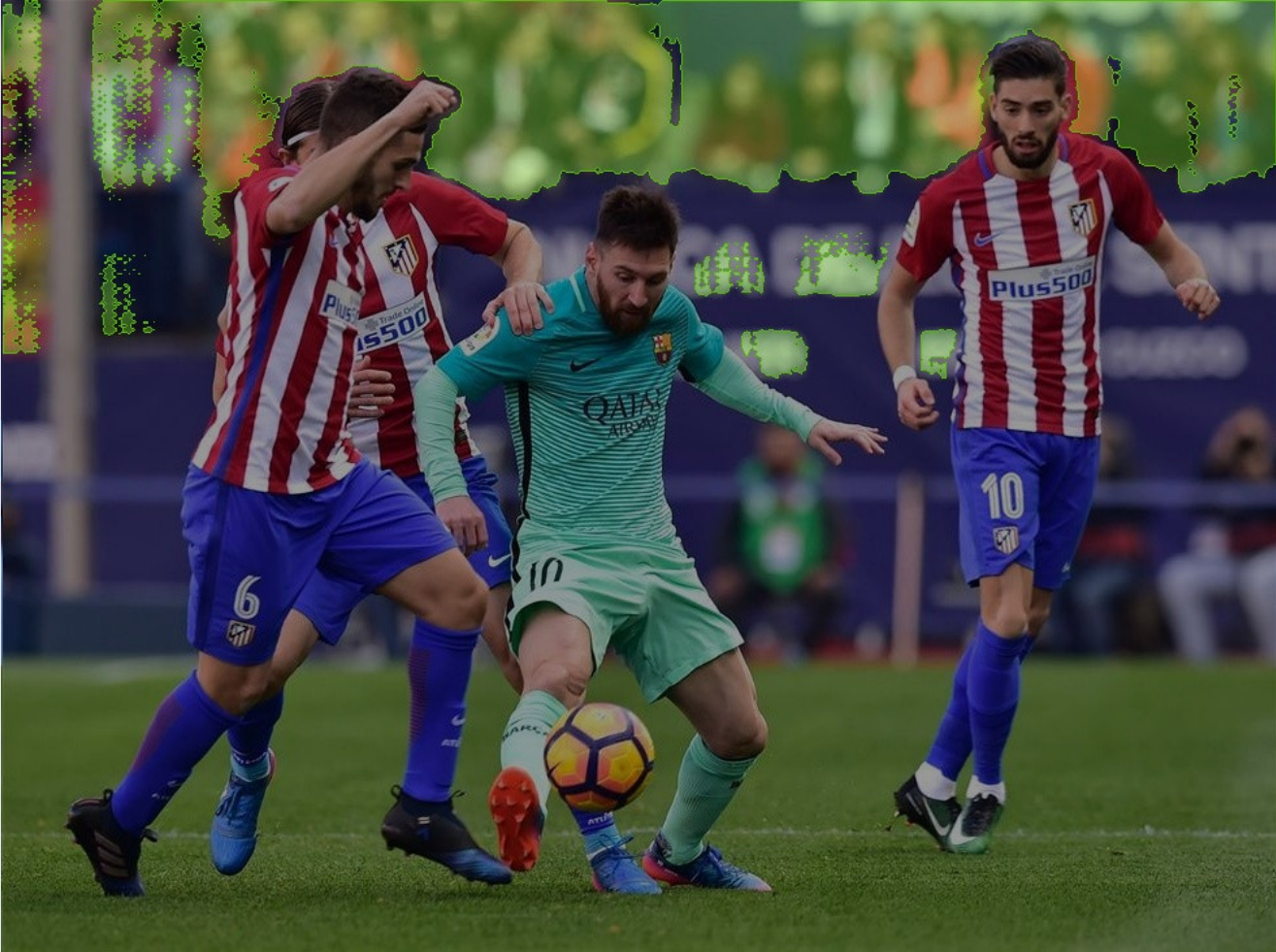}
    \end{subfigure}
    \caption{
    Multiple background instances may merge into a single mask when no bounding box is provided as a prompt.
    The yellow mask was refined and covers all spectators.
    Foreground instances are omitted in the left image for clarity.
    \\Left -- RTMDet \cite{RTMDet}, right -- BMP.
    }
    \label{fig:SUPPL-background}
\end{figure}

\textbf{Bounding box depending on max\_IoU}.
The last experiment (H1) involves prompting with a bounding box only for instances with $max\_IoU > 0.5$.
The rationale is that bounding boxes are typically accurate for isolated instances, where bounding box prompting improves results.
However, for highly overlapping instances, the bounding box is often inaccurate and degrades detection performance.
The results of this experiment are in \cref{tab:sam-ablation-study-other}.

As expected, the results fall between always prompting with bounding boxes and never using them.
While this approach significantly improves segmentation compared to prompting without bounding boxes, the improvement in detection over always prompting with bounding boxes is minor.
A qualitative analysis reveals that this method is primarily beneficial for low-resolution background instances, such as spectators in sports images. Without bounding box prompting, SAM often segments the entire background, leading to inaccuracies.
This phenomenon is not well captured in the evaluation, as background instances rarely have pose annotations and have limited detection and segmentation labels.
An example is shown in \cref{fig:SUPPL-background}.

\subsection{Summary}

The ablation study on automated SAM prompting is extensive and may seem overwhelming.
To provide a clear summary, the last rows of \cref{tab:sam-ablation-study-other} present two prompting methods used in BBox-Mask-Pose (BMP).

\textbf{D3}: This method is used in the BMP loop to balance refined masks with improved detection.
It primarily enhances detection accuracy while slightly improving segmentation.
Although it does not achieve the best standalone results, it performs best when used within the closed BMP loop with re-detections.

\textbf{J1}: This method is designed to refine masks and poses to produce high-quality estimates.
It is used, for instance, in BMP ablations (\cref{sec:results-ablation}) to loop SAM and MaskPose without re-detection.
It significantly improves segmentation and pose estimation but is not part of the reported BMP results.
J1 could be applied after the BMP loop terminates to further refine masks and bounding boxes, but we avoided this because it introduces additional overhead by requiring extra SAM (and possibly MaskPose) iterations.
While such micro-loops and adjustments could further improve the reported results, our focus is on maintaining clarity, showing that two simple loops are sufficient to improve detection, segmentation, and pose estimation.

\textbf{Pose estimation robustness}.
Pose estimation demonstrates notable robustness to the quality of estimated masks. MaskPose consistently produces accurate poses, even with low-quality masks (e.g., experiment C5 in \cref{tab:sam-ablation-study-num-neg-kpts}), and almost always outperforms the ViTPose \cite{ViTPose} baseline conditioned by the bounding box.
However, achieving the MaskPose-SAM-MaskPose self-improving loop requires employing several hand-crafted tweaks.
Among these, the Pose-Mask Consistency, as used in experiment J1 in \cref{tab:sam-ablation-study-other}, is particularly critical. 
Overall, BMP's pose estimation benefits more from refined detections and re-detection of background instances than from refining masks through SAM.
This highlights the importance of robust detection to improve overall performance within the BMP framework.

\section{Additional results}
\label{sec:additional-results}

\cref{tab:CIHP_results} shows results on CIHP dataset \cite{CIHP}.
BMP is the most effective in scenarios with max IoU between 0.5 and 1.0 (see also Tab. 3).
The improvement in non-crowd scenes (e.g. COCO) is negligible.
Note that not all crowd datasets are equal.
COCO, CIHP and CrowdPose feature group photos with many bboxes tightly squeezed next to each other.
On the other hand, OCHuman and part of CIHP feature entangled people with highly overlapping bboxes.
BMP excels in the most difficult scenes with overlapping bboxes, while not harming performance on group photos.

        

\begin{SCtable}[][bt]
    \centering
    \begin{tabular}{@{}l|l|l@{}}
        \toprule
        & det AP & mask AP \\
        \midrule
        RTMDet-l &  69.5   & 63.9    \\
        BMP $1\times$ &  69.4 \decr{-0.1}  & 65.7  \impr{+1.8}   \\ 
        BMP $2\times$ &  \textbf{69.7} \impr{+0.2}   & \textbf{65.9}  \impr{+2.0}    \\ 
        
        \bottomrule
    \end{tabular}
    \vspace{-1.0cm}
    \caption{
    \textbf{Detection results on CIHP \cite{CIHP}}.
    BMP brings a small improvement; CIHP is more similar to COCO than to OCHuman.  
    }
    \label{tab:CIHP_results}
\end{SCtable}

\section{Failure cases analysis}
\label{sec:failure-cases}

Here, we provide a detailed analysis of BMP failure cases.
While the most common issues are discussed in the paper, particularly in \cref{sec:conclusions,fig:BMP-failures}, this section offers additional examples and introduces a previously unmentioned type of error, instance merging.

\begin{figure}[tb]
    \centering
    \begin{subfigure}{0.302\linewidth}
        \centering
        \includegraphics[width=\textwidth,page=27]{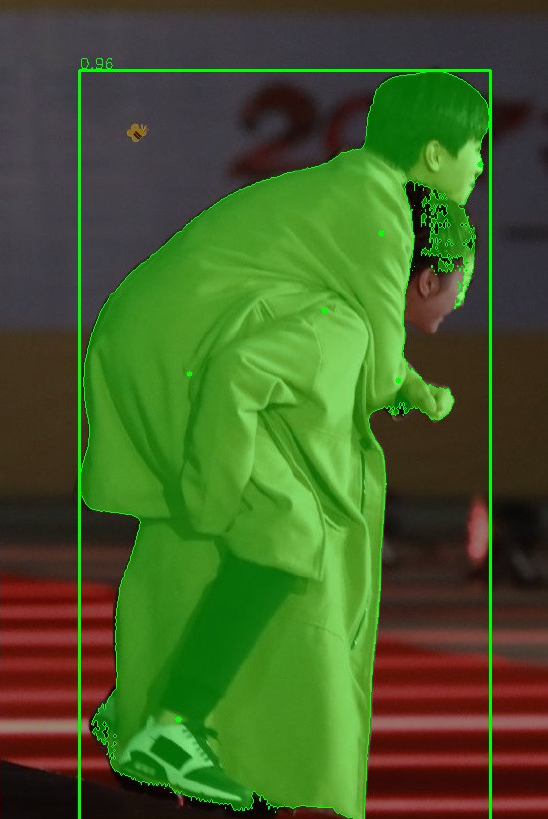}
        \caption{
        Two people in matching coats.
        \\
        }
    \end{subfigure}
    \hfill
    \begin{subfigure}{0.32\linewidth}
        \centering
        \includegraphics[width=\textwidth,page=47]{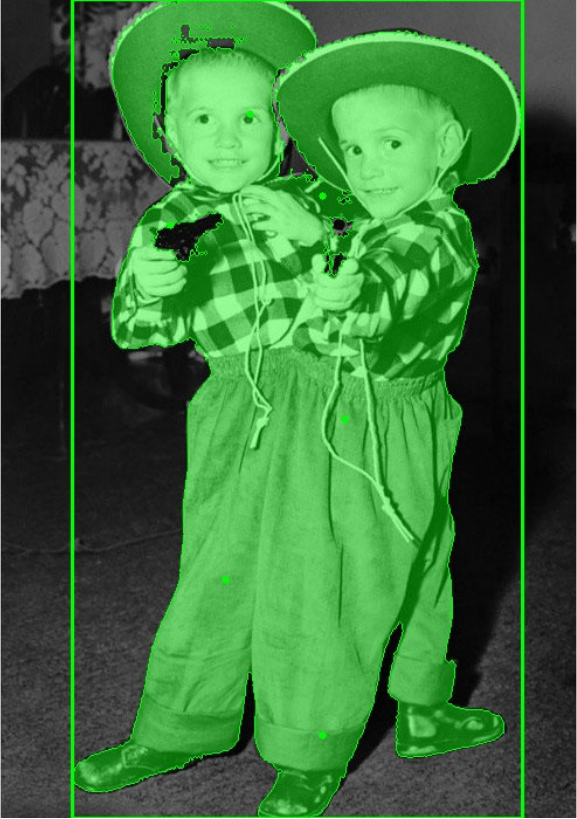}
        \caption{
        Two boys in one pair of pants, wearing matching shirts.
        }
    \end{subfigure}
    \hfill
    \begin{subfigure}{0.265\linewidth}
        \centering
        \includegraphics[width=\textwidth,page=28]{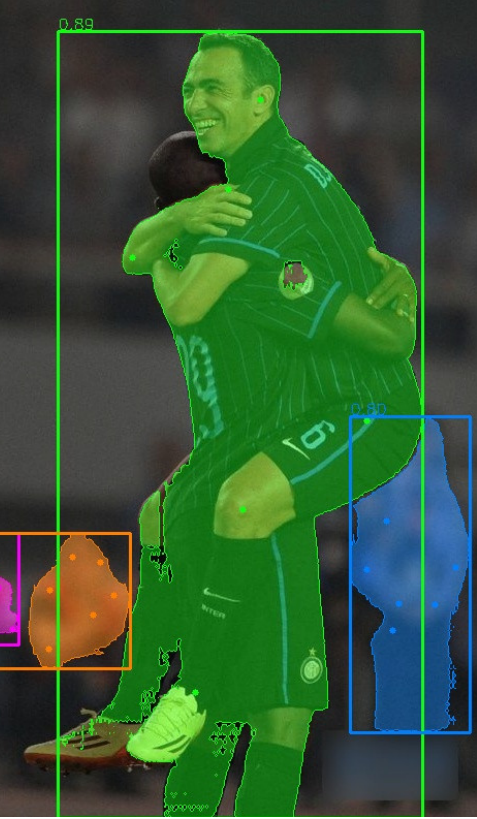}
        \caption{
        Two players with matching jerseys.
        }
    \end{subfigure}
    \caption{
    Instances not split even after mask refinement by SAM \cite{SAM2}, typically due to similar or identical textures.
    }
    \label{fig:SUPPL-merged}
\end{figure}

\textbf{Merging instances}.
Even though BMP is designed to decouple instances merged by the detector, and MaskPose performs well in such cases, SAM can mistakenly merge instances if it is incorrectly prompted or if the instances have similar textures.
Prominent examples of these failures are shown in \cref{fig:SUPPL-merged}. 

BMP struggles to address these issues because bounding box prompting would also fail, given that the detected bounding box already merges the instances.
Furthermore, Pose-Mask Consistency (P-Mc) does not help in such cases, as only one instance is detected.
Without negative keypoints, a large mask that merges multiple instances (or even covers the entire image) would still achieve $P-Mc = 1.0$, since all positive keypoints fall within the mask and no negative keypoints are present to penalize the score.

\begin{figure}[tb]
    \centering
    \begin{subfigure}{0.297\linewidth}
        \centering
        \includegraphics[width=\textwidth,page=7]{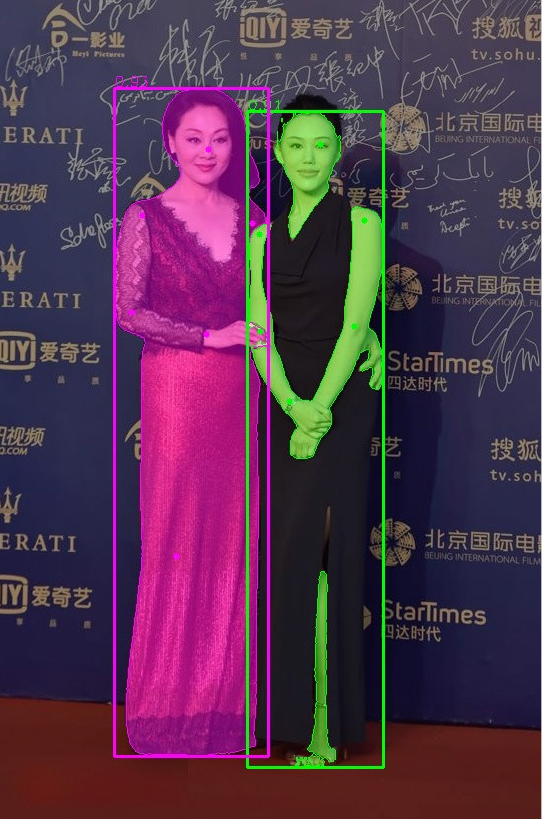}
        \caption{
        }
    \end{subfigure}
    \hfill
    \begin{subfigure}{0.35\linewidth}
        \centering
        \includegraphics[width=\textwidth,page=9]{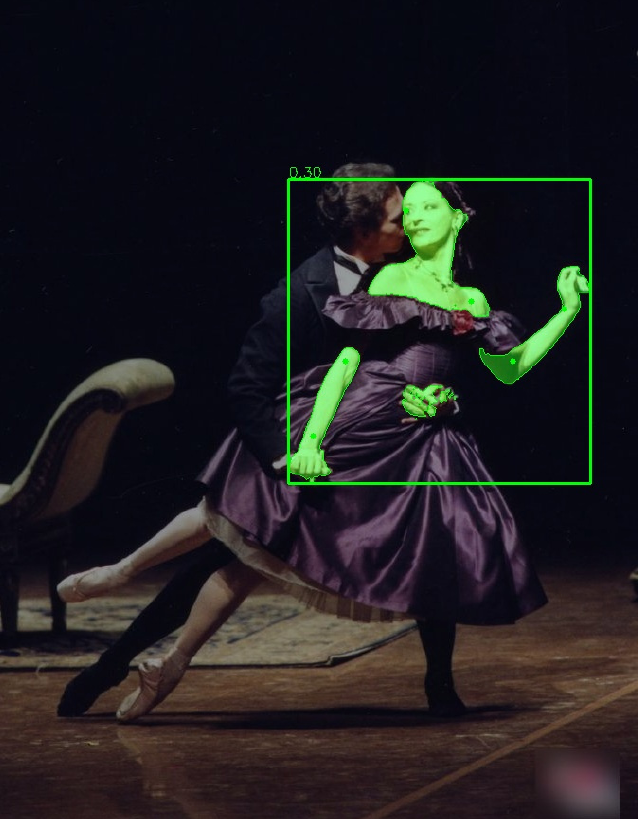}
        \caption{
        }
    \end{subfigure}
    \hfill
    \begin{subfigure}{0.282\linewidth}
        \centering
        \includegraphics[width=\textwidth,page=30]{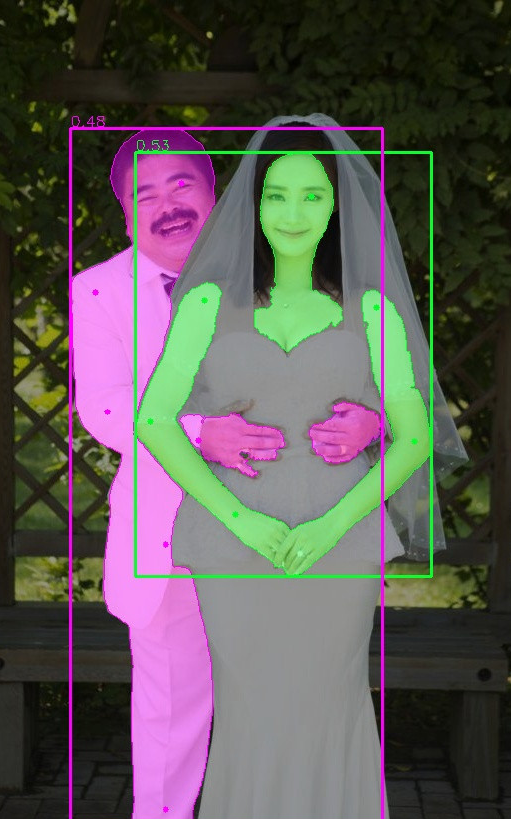}
        \caption{
        }
    \end{subfigure}
    \caption{
    Oversegmentation.
    Green instances have incorrect masks -- only the skin is segmented, excluding the clothes.
    This issue commonly occurs with clothing that exposes bare shoulders, such as dresses or jerseys.
    Keypoints used for SAM prompting are marked (best viewed in zoom).
    }
    \label{fig:SUPPL-clothes}
\end{figure}

\textbf{Segmenting clothes} instead of the whole person.
This issue, illustrated in \cref{fig:SUPPL-clothes}, is particularly common in OCHuman, where many individuals wear specific clothing.
The problem frequently arises when a person has bare shoulders, such as in an evening dress or basketball jersey.
In such cases, shoulder, facial, knee, elbow, and wrist keypoints, which are on the skin rather than clothing, prompt SAM to segment only the skin, leaving the clothing unsegmented.
Hip and sometimes ankle keypoints could help refine segmentation, but these are typically low-confidence predictions and are often not selected.

Unsegmented clothing causes downstream issues as the masking-out step leaves the clothes visible.
In subsequent BMP iterations, the detector identifies these as separate instances, as shown in \cref{fig:BMP-failures}. 

We suggest two potential solutions. The first is to improve SAM prompting to include clothing in the segmentation.
The bounding box prompt could address this specific case, but it hinders performance in other scenarios, as detailed in \cref{fig:SAM-ablation-bbox} and \cref{sec:prompting-sam}.
The second is to fine-tune the detector to ignore clothing when the skin is masked out.
However, this approach risks reducing the detector's generalizability and causing overfitting to scenarios with visible skin and faces, which we believe is not a viable long-term solution.

\begin{figure}[tb]
    \centering
    \begin{subfigure}{0.49\linewidth}
        \centering
        \includegraphics[width=\textwidth,page=33]{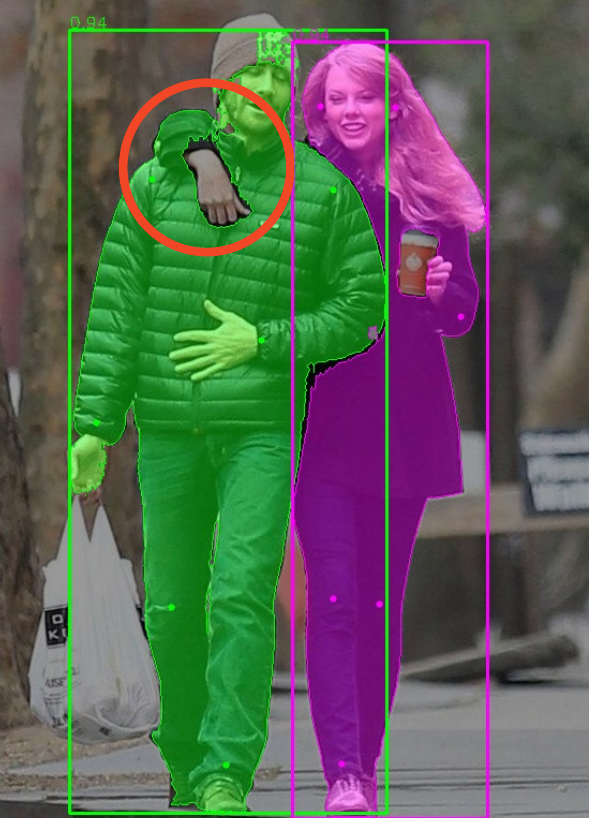}
    \end{subfigure}
    \hfill
    \begin{subfigure}{0.49\linewidth}
        \centering
        \includegraphics[width=\textwidth,page=34]{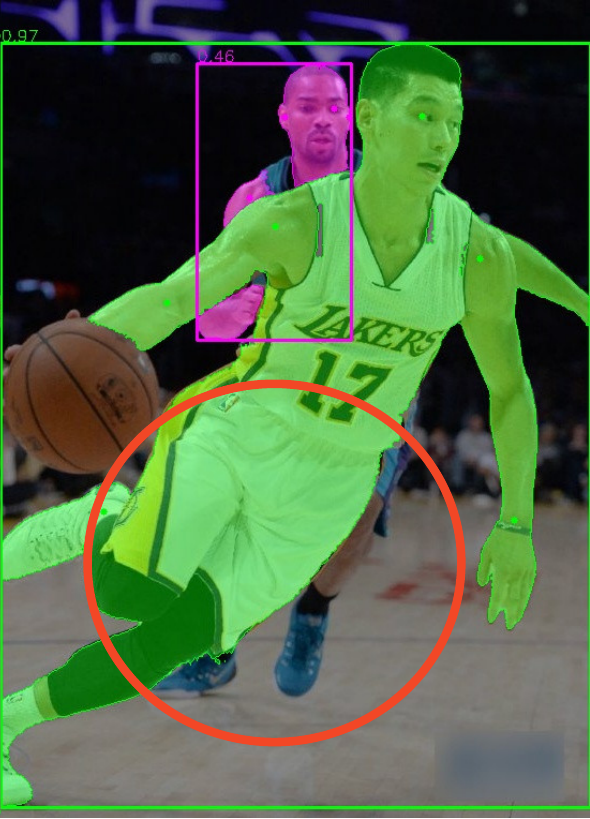}
    \end{subfigure}
    \caption{
    Images where SAM \cite{SAM2} successfully decoupled instances but failed to segment a disconnected body part.
    These parts remain unmasked and risk being re-detected, as illustrated in \cref{fig:BMP-failures-missed-limb}.
    Keypoints used for SAM prompting are marked (best viewed in zoom).
    }
    \label{fig:SUPPL-missed-limb}
\end{figure}

\textbf{Missing body parts}.
When SAM fails to segment a body part, it remains unmasked and may be redetected in the next stage, as shown in \cref{fig:BMP-failures,fig:SUPPL-missed-limb}.
This issue is even more pronounced when prompting with a bounding box, as detected bounding boxes often exclude disconnected limbs, leaving SAM unable to recover them. For this reason, we avoid prompting with the bounding box in the BMP loop.

Missed limbs could potentially be addressed by better alignment between pose and mask.
If the refined mask is inconsistent with the prompted pose, SAM could be restarted with different prompts to minimize missed limbs.
However, if the limb is also missed by MaskPose, BMP cannot resolve the issue.

\begin{figure}[tb]
    \centering
    \begin{subfigure}{\linewidth}
        \centering
        \includegraphics[width=0.49\textwidth,page=40]{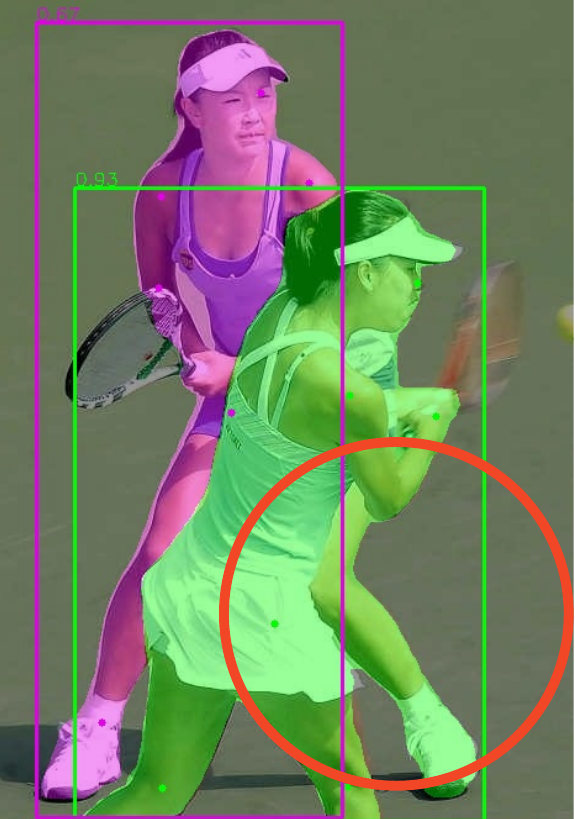}
        \hfill
        \includegraphics[width=0.49\textwidth,page=39]{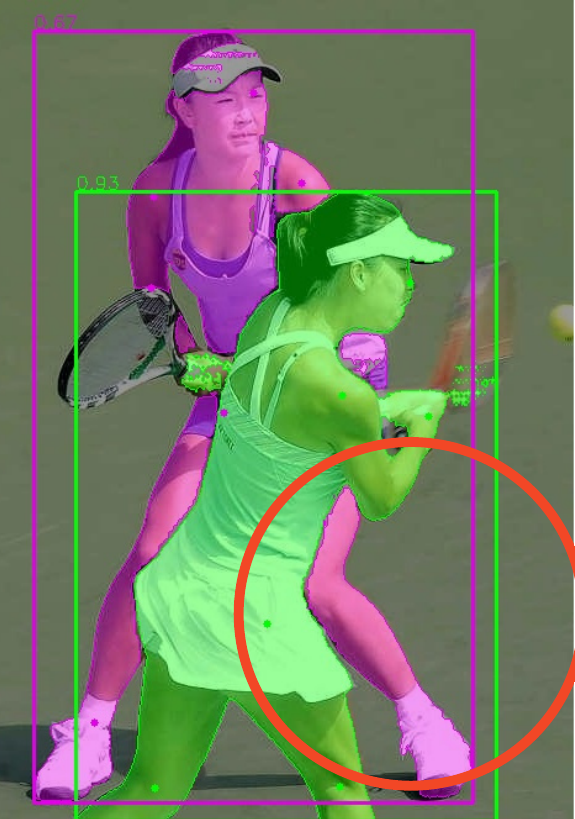}
    \end{subfigure}
    
    \begin{subfigure}{\linewidth}
        \centering
        \includegraphics[width=0.49\textwidth,page=44]{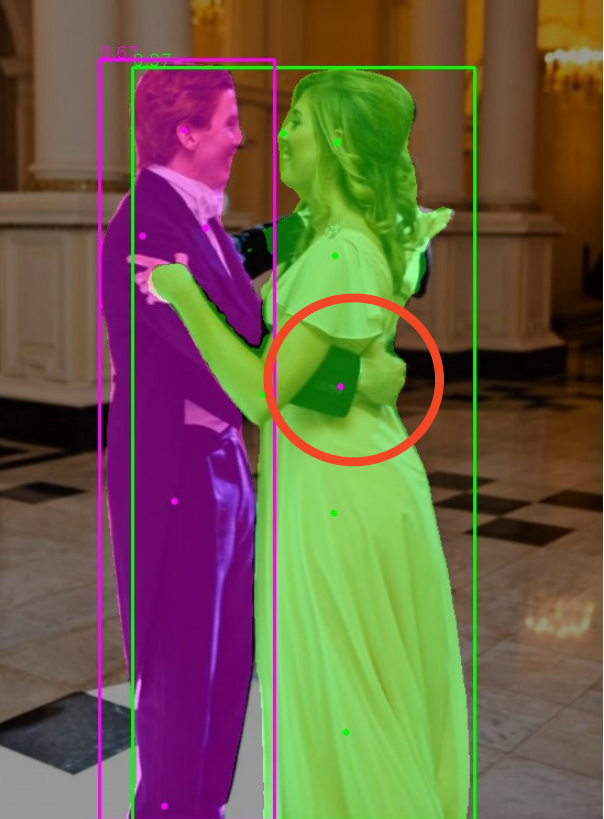}
        \hfill
        \includegraphics[width=0.49\textwidth,page=43]{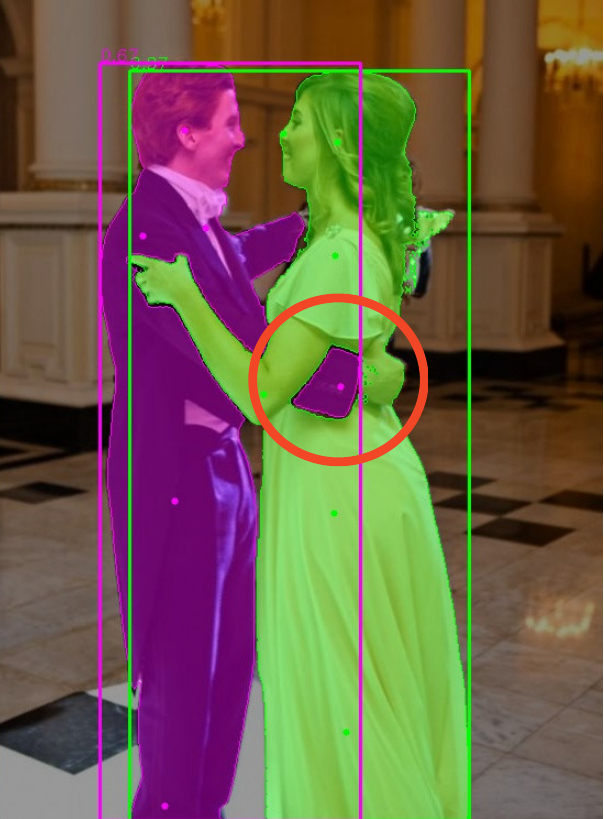}
    \end{subfigure}
    
    \caption{
    Images where BMP improves detection and segmentation using its pose estimates and SAM prompting with selected keypoint.
    Bounding box prompting did not lead to comparable results. 
    Keypoints used for SAM prompting are marked (best viewed in zoom).
    Left -- RTMDet \cite{RTMDet}, right -- BMP.
    \vspace{28em}
    }
    \label{fig:SUPPL-good-fix}
\end{figure}

\begin{figure}[tb]
    \centering
    
    \begin{subfigure}{\linewidth}
        \centering
        \includegraphics[width=0.49\textwidth,page=42]{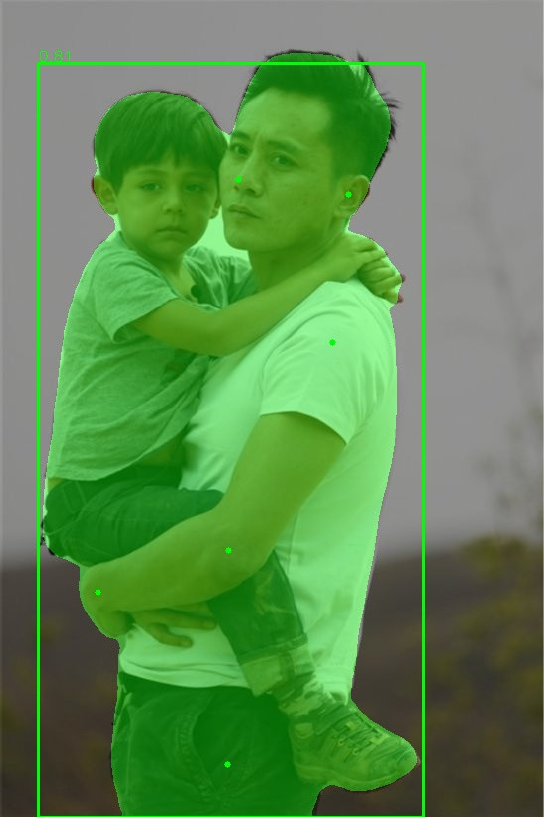}
        \hfill
        \includegraphics[width=0.49\textwidth,page=41]{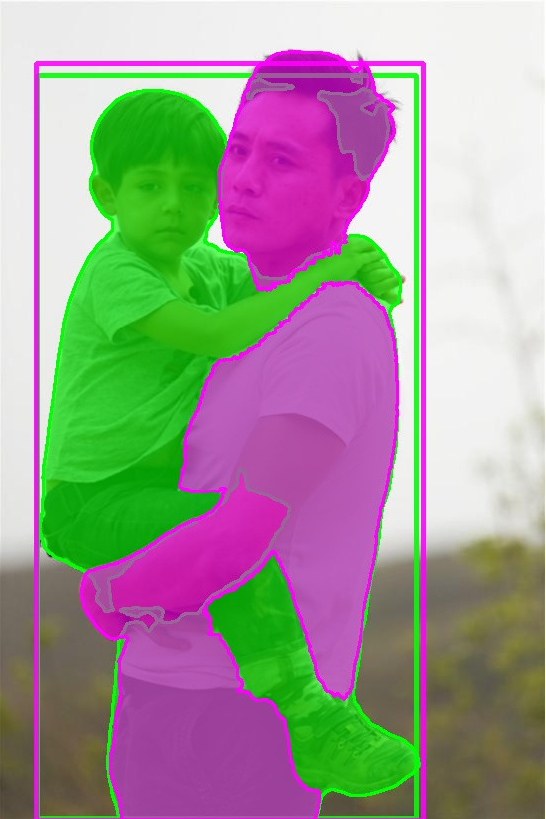}
    \end{subfigure}
    
    \begin{subfigure}{\linewidth}
        \centering
        \includegraphics[width=0.49\textwidth,page=38]{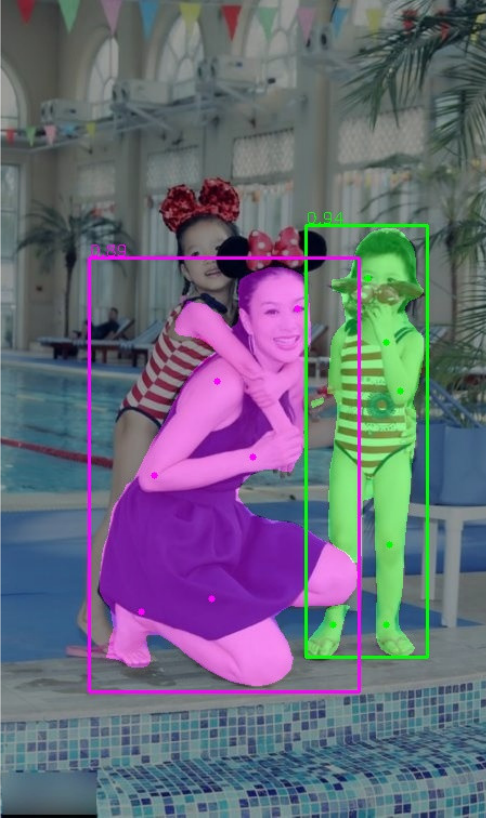}
        \hfill
        \includegraphics[width=0.49\textwidth,page=37]{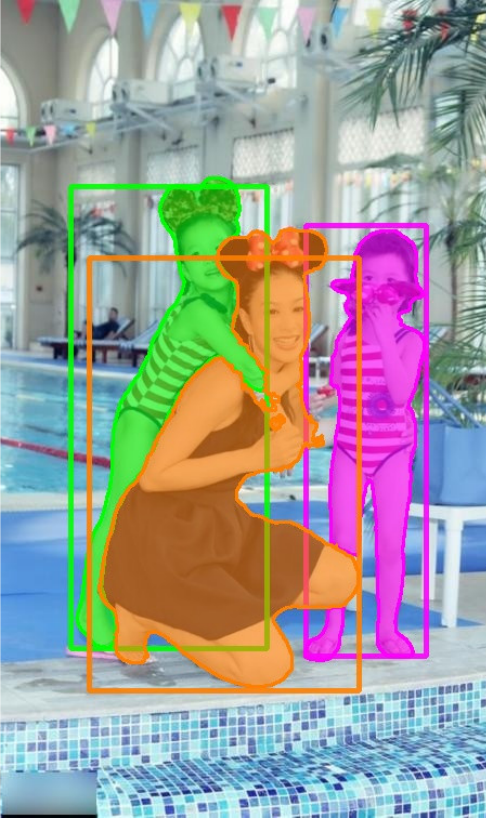}
    \end{subfigure}
    
    \begin{subfigure}{\linewidth}
        \centering
        \includegraphics[width=0.49\textwidth,page=45]{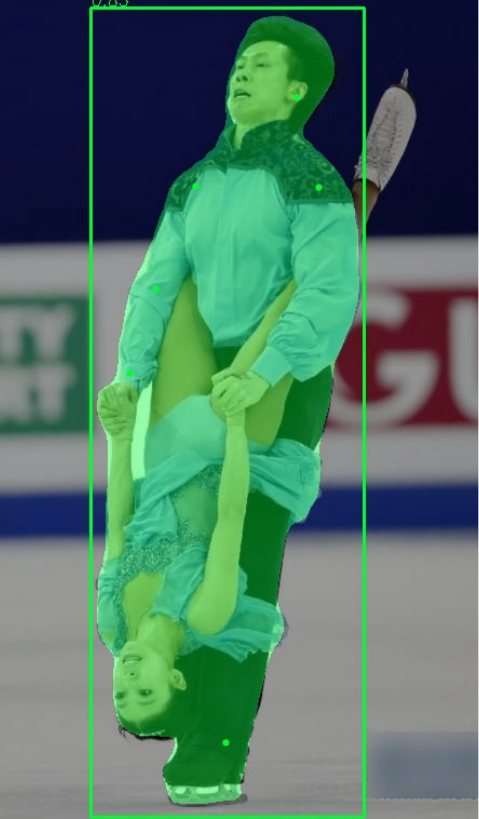}
        \hfill
        \includegraphics[width=0.49\textwidth,page=46]{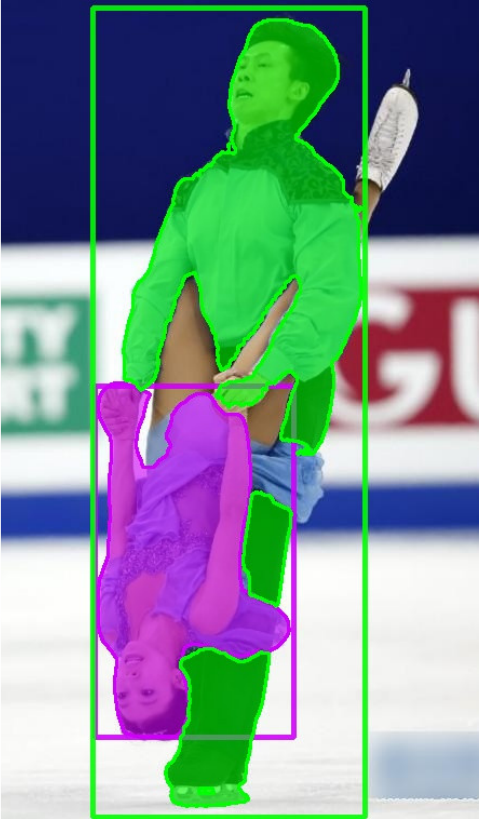}
    \end{subfigure}
    
    \caption{
    Two iterations of BMP successfully decouple merged instances, even in challenging images with upside-down people.
    \\Left -- RTMDet \cite{RTMDet}, right -- BMP.
    \vspace{5em}
    }
    \label{fig:SUPPL-good-merge}
\end{figure}

\textbf{Correct examples}.
BMP performs reliably in most cases, as demonstrated by the quantitative results.
\cref{fig:SUPPL-good-fix,fig:SUPPL-good-merge} showcase examples of successful detection and segmentation in challenging multi-body scenarios, including cases where a person is upside down.

In particular, \cref{fig:SUPPL-good-fix} highlights the ability of BMP to balance segmentation and detection, as discussed in \cref{fig:SUPPL-det-vs-segm}.
The improvements are significant, with more precise segmentation and accurate instance counts in the scene.
Some small body parts may occasionally be assigned to the wrong instance, but overall performance remains strong.

\section{Additional ablation Study}
\label{sec:suppl-ablation}

\subsection{Semi-transparency for MaskPose}

\cref{fig:alpha-levels} shows preliminary experiments on the $\alpha$ values in MaskPose.
When $\alpha = 0$, the model loses the background context and becomes sensitive to detected mask quality.
For $\alpha \in [0.2, 0.8]$, the model combines the foreground and the background and exhibits good and stable performance.

\begin{figure}[bt]
  \centering
  \begin{tikzpicture}
    \begin{axis}[
      xmin=-0.02, xmax=1.02,
      ymin=38.3, ymax=42.3,
      xtick={0,0.2,0.5,0.8,1},   
      xlabel={$\alpha$},
      ylabel={OCHuman test AP},
      grid=both,              
      major grid style={line width=0.4pt, draw=gray!50},
      minor grid style={line width=0.2pt, draw=gray!20},
      width=\linewidth, height=4cm
    ]
  

        \draw[
          fill=red,
          fill opacity=0.2,
          draw=none
        ] 
          (axis cs:0.17,42.2) rectangle (axis cs:0.83,41.1)
          ;
      
      \addplot[
        BrightBlue,
        mark=x,
        mark size=3pt,               
        mark options={thick},        
        line width=0.8pt
      ] coordinates {
        (0.0, 38.56)
        (0.2, 41.70)
        (0.5, 41.87) 
        (0.8, 41.62)
        (1.0, 38.90)
      };
      \draw[->, CobaltBlue, line width=1.2pt
      ] 
        (axis cs:0.78,38.9) 
        node[left] {No mask = ViTPose}
        --
        (axis cs: 0.98,38.9) 
        ;
      
      \draw[->, CobaltBlue, line width=1.2pt] 
        (axis cs:0.22,39.8) 
        node[right] {No background}
        --
        (axis cs: 0.02,38.5) 
        ;
    \node[
      text=red,
      anchor=center
    ] at (axis cs:0.5,41.35) {Same performance};

    \end{axis}
  \end{tikzpicture}

  \caption{
  MaskPose performance with \textbf{different values of $\alpha$}.
  Fine-tuning for 5 epochs on 10\% of dataset, masks detected.
  }
  \label{fig:alpha-levels}
\end{figure}

\subsection{Number of parameters of BMP}

\begin{table}[tb]
    \centering
    \begin{tabular}{@{}c c c c |r r r}
        \toprule
        pose   & SAM    & pose   & loops      & bbox  & pose & params \\
        \midrule
        \cmark & \cmark & \xmark & $1\times$  & 31.1     & 45.3    & 225 M \\
        \cmark & \cmark & \xmark & $2\times$  & \textbf{32.1}     & \textbf{48.6}    & 369 M \\
        \midrule
        \cmark & \cmark & \textcolor{magenta}{\cmark} & $1\times$  & 31.1     & 46.4    & 312 M \\
        \textcolor{magenta}{\xmark} & \cmark & \xmark & $2\times$  & \underline{31.9}     & \underline{47.3}    & 282 M \\
        \cmark & \textcolor{magenta}{\xmark} & \xmark & $2\times$  & 30.8     & 47.0    & 201 M \\
        \bottomrule
    \end{tabular}
    \caption{
    \textbf{Ablation study} of BBox-Mask-Pose components evaluated on OCHuman-val.
    Bbox and pose evaluated with AP.
    The sum of trainable parameters approximates computational complexity.
    First row corresponds to BMP $1\times$, second to BMP $2\times$.
    }
    \label{tab:ablation-study-params}
\end{table}

\cref{tab:ablation-study} in \cref{sec:results-ablation} shows the performance change with and without various BMP components.
For clarity, we also present \cref{tab:ablation-study-params}, which shows the same result along with the number of trainable parameters of the whole loop.
For example, combining the detector (RTMDet-l) with 57M parameters and the pose model (ViTPose-b) with 87M parameters results in 144M trainable parameters.

Omitting SAM from the loop significantly reduces parameters, but also sharply decreases performance.
Running the pose estimation again after the SAM refinement increases parameter usage by 40\%, from 225M to 312M.

\begin{figure}[tb]
    \centering
    
    \begin{subfigure}{\linewidth}
        \centering
        \includegraphics[width=\textwidth]{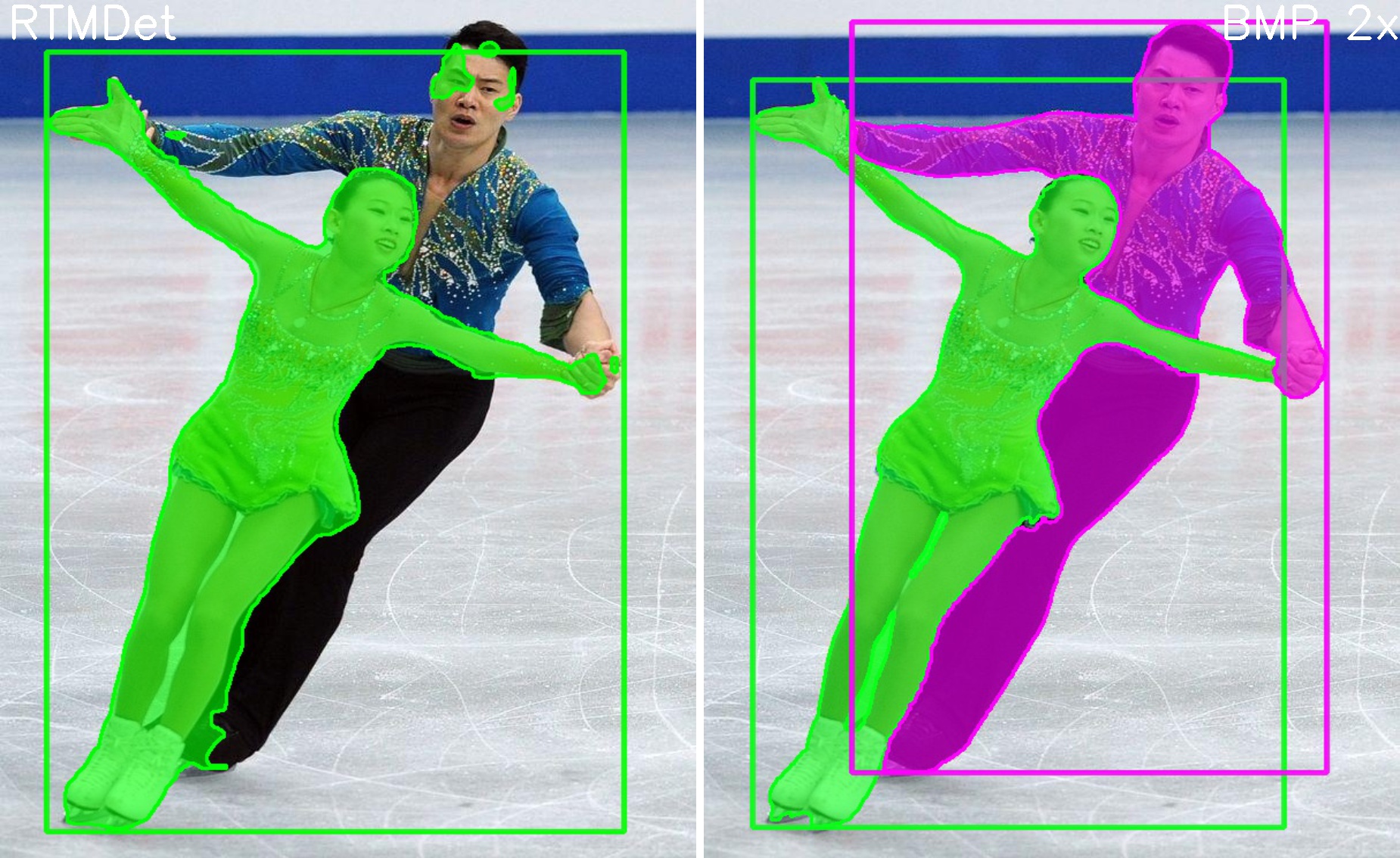}
    \end{subfigure}

    \begin{subfigure}{\linewidth}
        \centering
        \includegraphics[width=\textwidth]{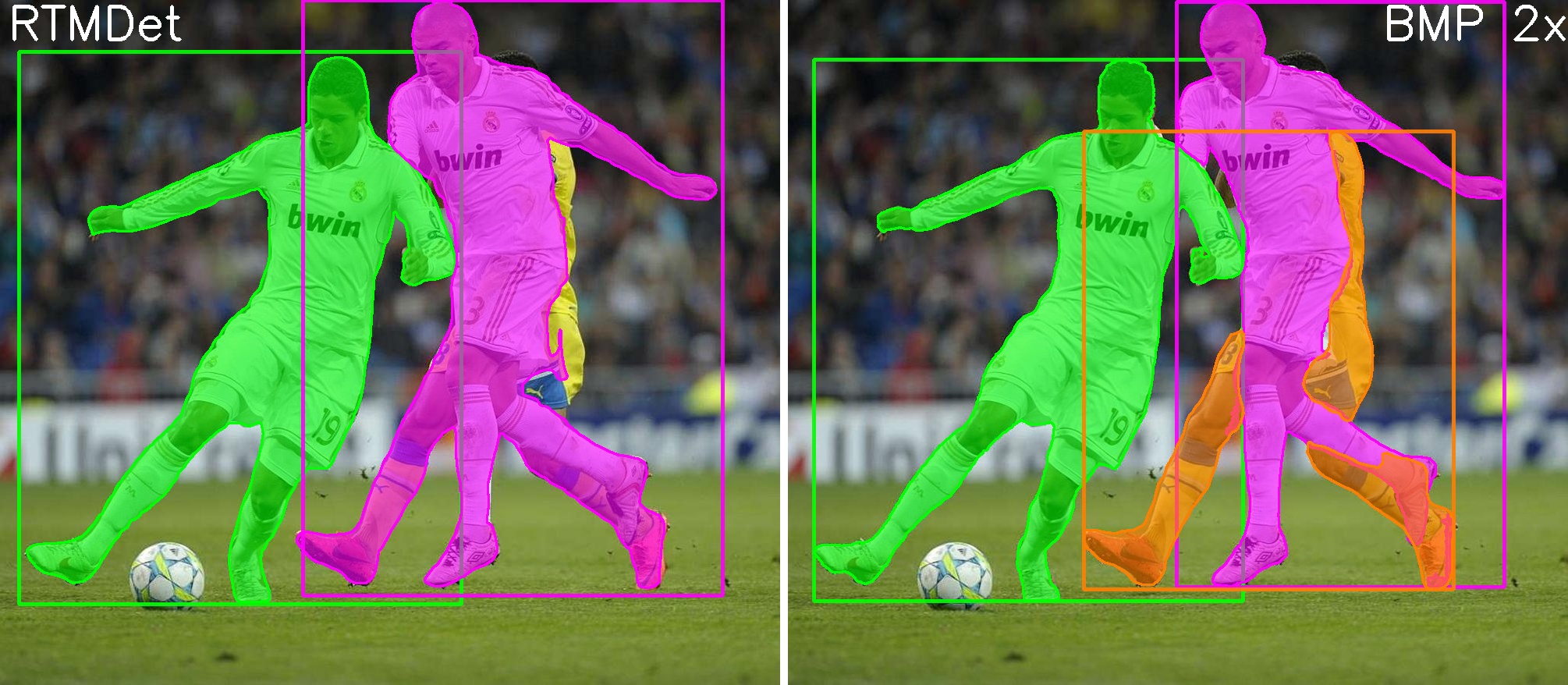}
    \end{subfigure}

    \begin{subfigure}{\linewidth}
        \centering
        \includegraphics[width=\textwidth]{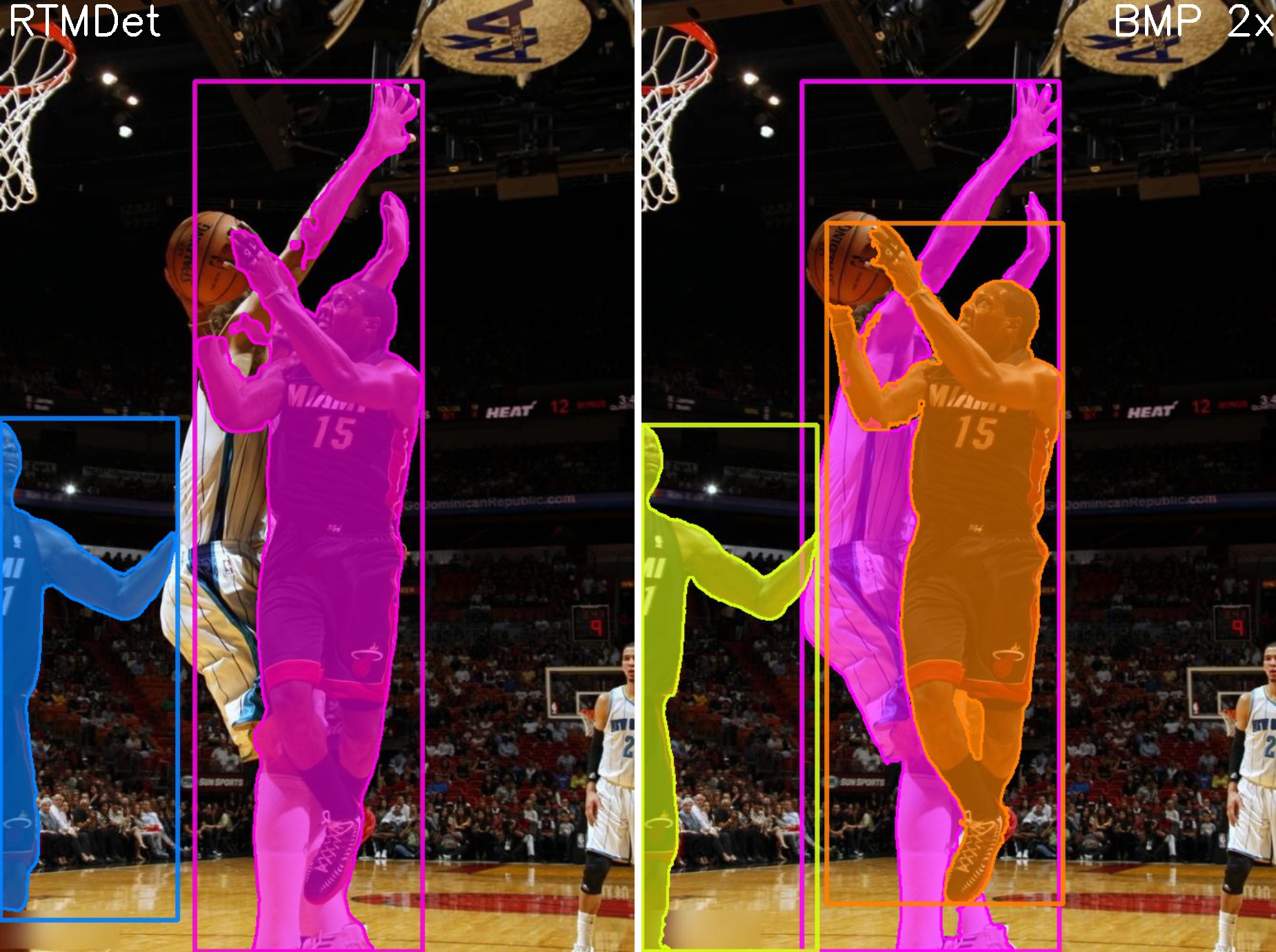}
    \end{subfigure}
    
    \begin{subfigure}{\linewidth}
        \centering
        \includegraphics[width=\textwidth]{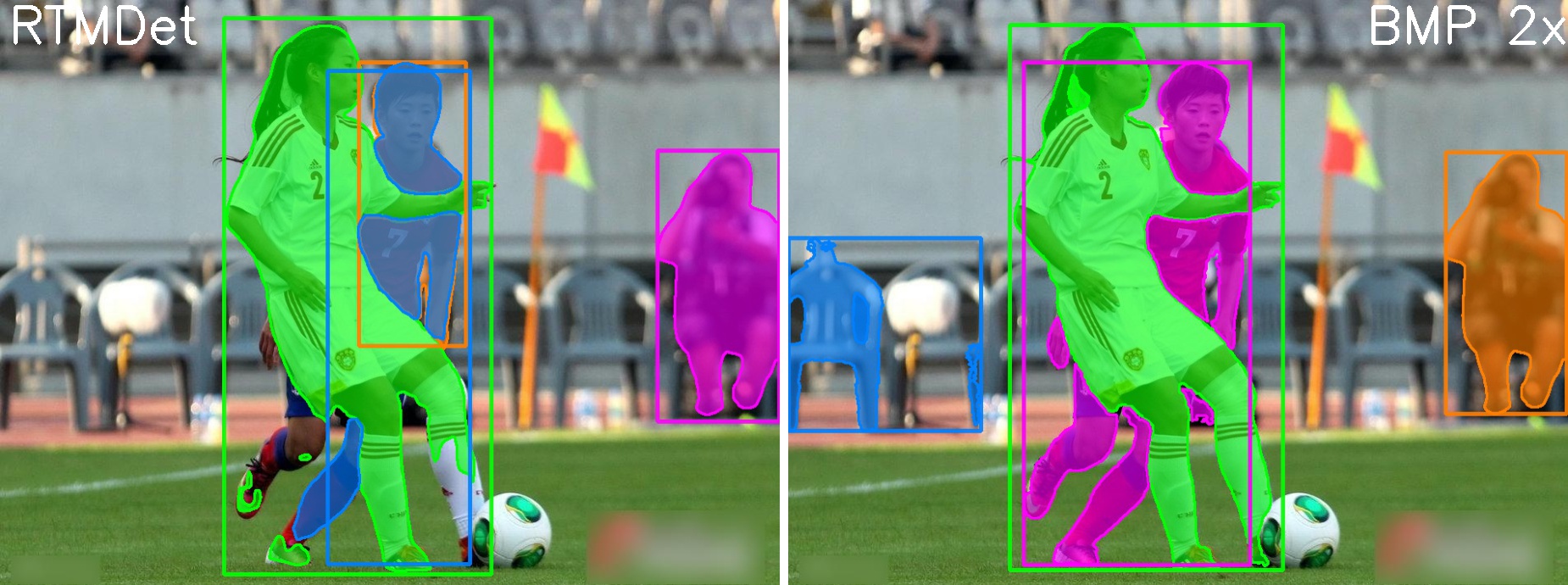}
    \end{subfigure}
    
    \begin{subfigure}{\linewidth}
        \centering
        \includegraphics[width=\textwidth]{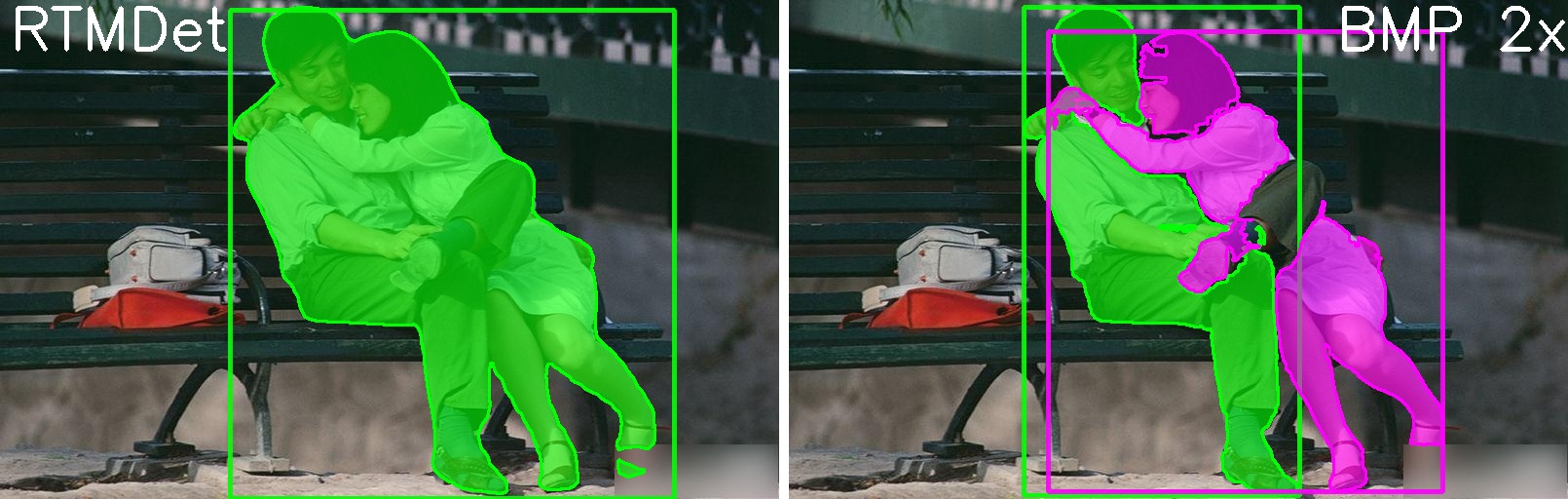}
    \end{subfigure}
    
    \caption{
    Qualitative results on the OCHuman dataset.
    \\Left -- RTMDet \cite{RTMDet}, right -- BMP $2\times$.
    \vspace{5em}
    }
    \label{fig:SUPPL-results}
\end{figure}

\begin{figure}[tb]
    \centering
    
    \begin{subfigure}{\linewidth}
        \centering
        \includegraphics[width=\textwidth]{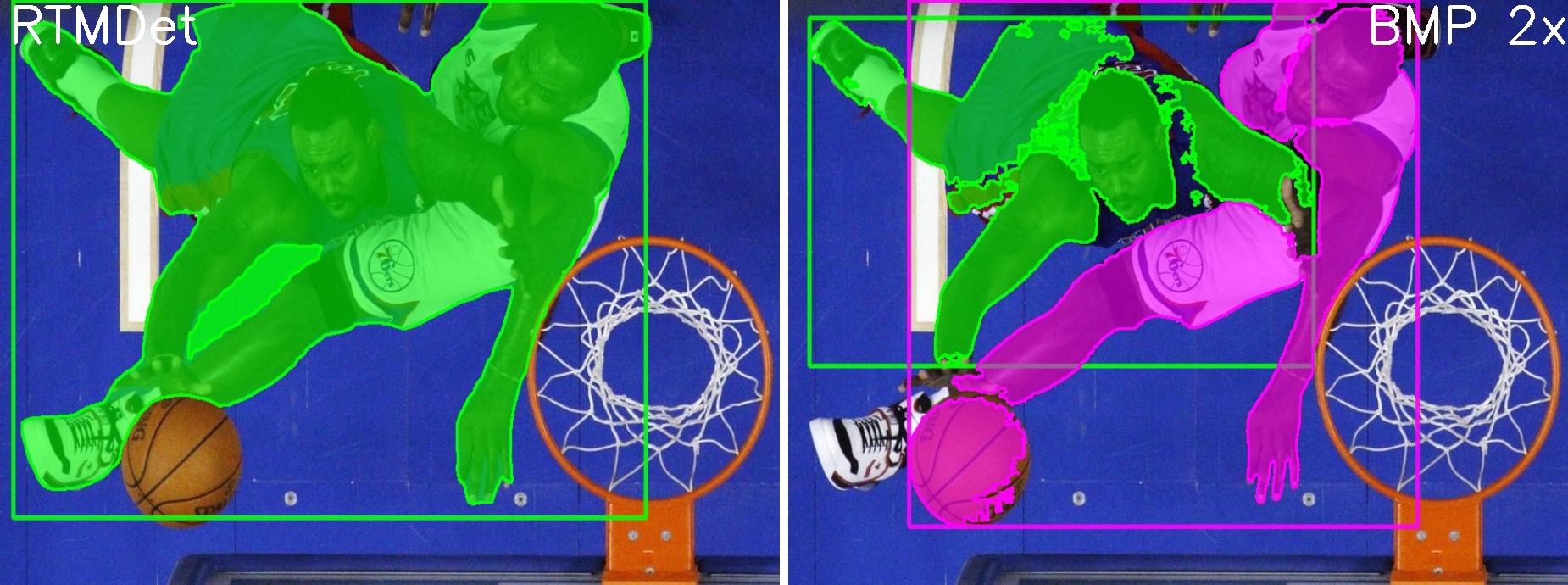}
    \end{subfigure}

    \begin{subfigure}{\linewidth}
        \centering
        \includegraphics[width=\textwidth]{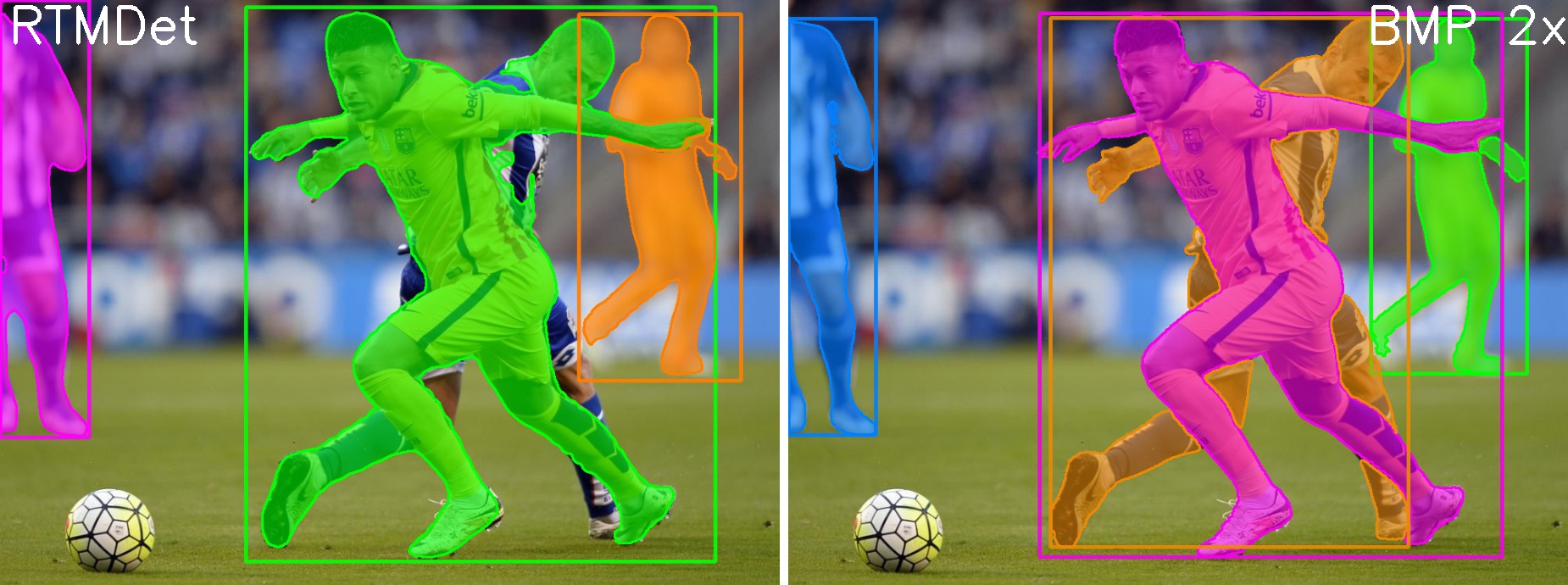} 
    \end{subfigure}

    \begin{subfigure}{\linewidth}
        \centering
        \includegraphics[width=\textwidth]{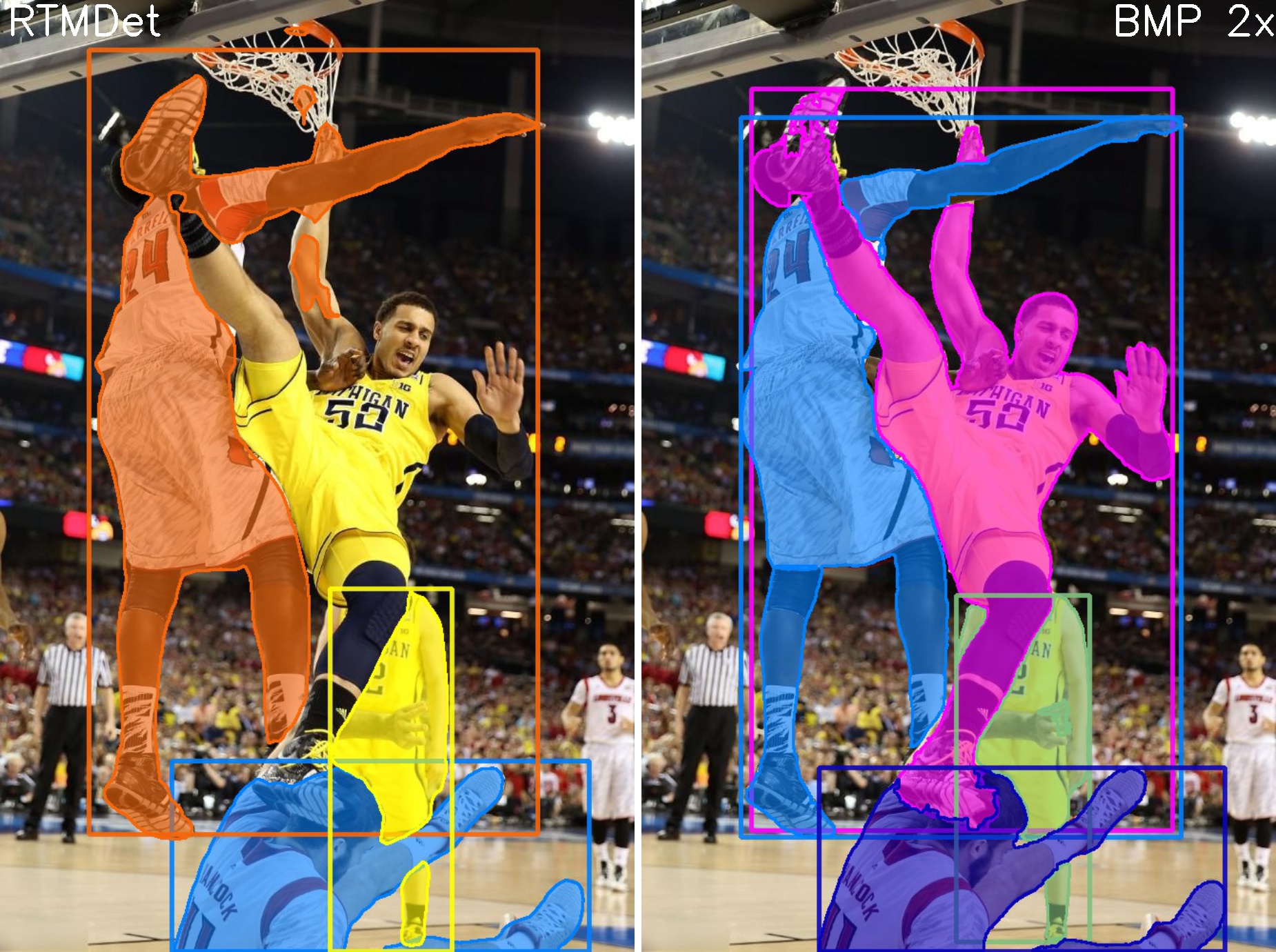}
    \end{subfigure}

    \begin{subfigure}{\linewidth}
        \centering
        \includegraphics[width=\textwidth]{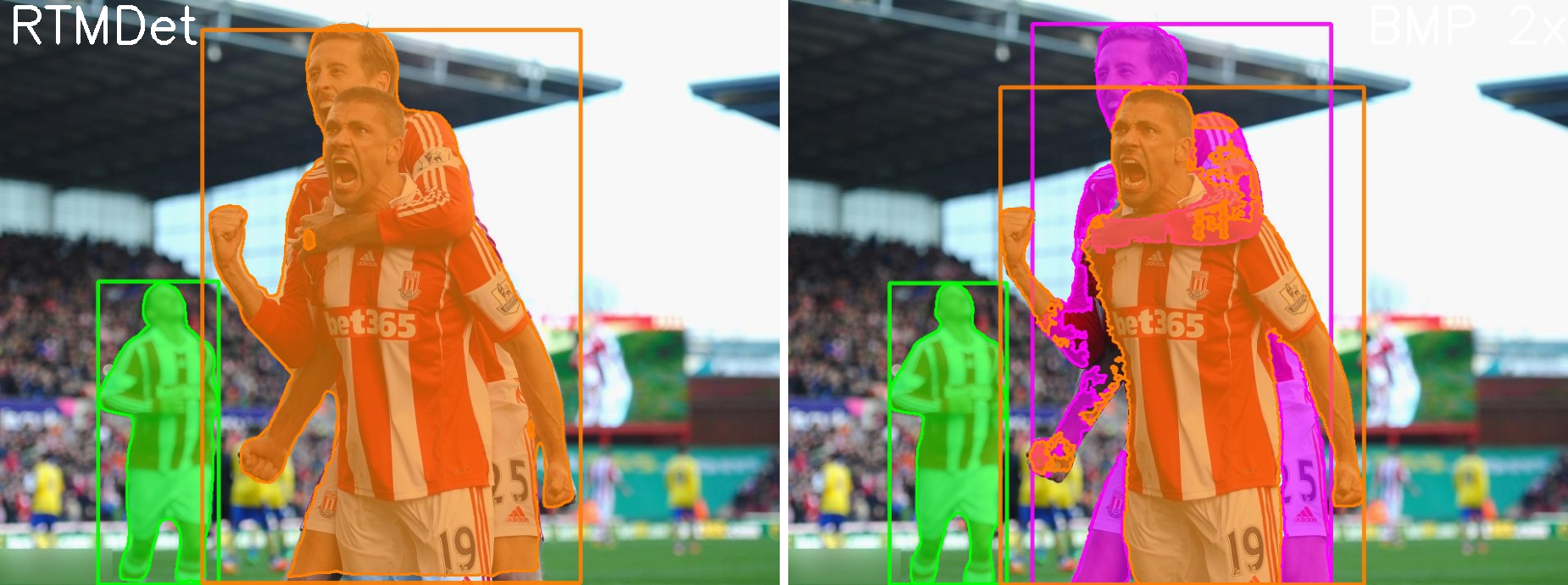}
    \end{subfigure}
    
    \begin{subfigure}{\linewidth}
        \centering
        \includegraphics[width=\textwidth]{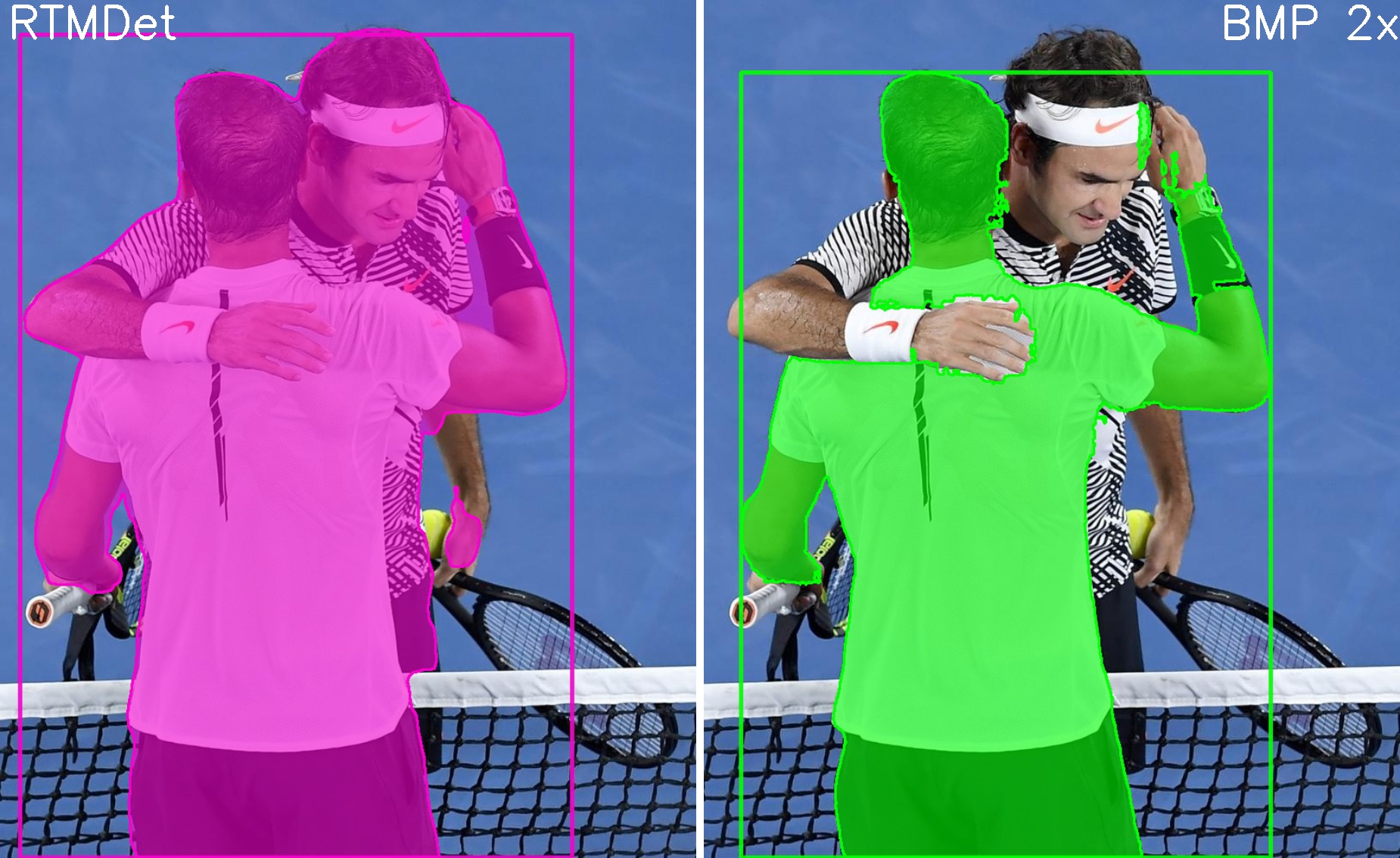}
    \end{subfigure}
    
    \caption{
    More qualitative results on the OCHuman dataset.
    \\Left -- RTMDet \cite{RTMDet}, right -- BMP $2\times$.
    \vspace{5em}
    }
    \label{fig:SUPPL-results2}
\end{figure}

%% file: main.bbl
\begin{thebibliography}{43}
\providecommand{\natexlab}[1]{#1}
\providecommand{\url}[1]{\texttt{#1}}
\expandafter\ifx\csname urlstyle\endcsname\relax
  \providecommand{\doi}[1]{doi: #1}\else
  \providecommand{\doi}{doi: \begingroup \urlstyle{rm}\Url}\fi

\bibitem[Ahmad et~al.(2022{\natexlab{a}})Ahmad, Khan, Kim, and Lee]{MultiPoseSeg}
Niaz Ahmad, Jawad Khan, Jeremy~Yuhyun Kim, and Youngmoon Lee.
\newblock Multiposeseg: Feedback knowledge transfer for multi-person pose estimation and instance segmentation.
\newblock \emph{2022 26th International Conference on Pattern Recognition (ICPR)}, pages 2086--2092, 2022{\natexlab{a}}.

\bibitem[Ahmad et~al.(2022{\natexlab{b}})Ahmad, Khan, Kim, and Lee]{PosePlusSeg}
Niaz Ahmad, Jawad Khan, Jeremy~Yuhyun Kim, and Youngmoon Lee.
\newblock Joint human pose estimation and instance segmentation with poseplusseg.
\newblock In \emph{AAAI Conference on Artificial Intelligence}, 2022{\natexlab{b}}.

\bibitem[Andriluka et~al.(2014)Andriluka, Pishchulin, Gehler, and Schiele]{MPII}
Mykhaylo Andriluka, Leonid Pishchulin, Peter Gehler, and Bernt Schiele.
\newblock 2d human pose estimation: New benchmark and state of the art analysis.
\newblock In \emph{IEEE Conference on Computer Vision and Pattern Recognition (CVPR)}, 2014.

\bibitem[Arthur and Vassilvitskii(2006)]{kmeans++}
David Arthur and Sergei Vassilvitskii.
\newblock k-means++: The advantages of careful seeding.
\newblock Technical report, Stanford, 2006.

\bibitem[Azarian et~al.(2022)Azarian, Das, Park, and Porikli]{PoseSegTTA}
Kambiz Azarian, Debasmit Das, Hyojin Park, and Fatih~Murat Porikli.
\newblock Test-time adaptation vs. training-time generalization: A case study in human instance segmentation using keypoints estimation.
\newblock \emph{2023 IEEE/CVF Winter Conference on Applications of Computer Vision Workshops (WACVW)}, pages 411--420, 2022.

\bibitem[Cai et~al.(2024)Cai, Gao, Zheng, Zhou, and Huang]{CrowdSAM}
Zhi Cai, Yingjie Gao, Yaoyan Zheng, Nan Zhou, and Di Huang.
\newblock Crowd-sam: Sam as a smart annotator for object detection in crowded scenes.
\newblock In \emph{Proceedings of the European Conference on Computer Vision (ECCV)}, 2024.

\bibitem[Cao et~al.(2017)Cao, Simon, Wei, and Sheikh]{OpenPose}
Zhe Cao, Tomas Simon, Shih-En Wei, and Yaser Sheikh.
\newblock Realtime multi-person 2d pose estimation using part affinity fields.
\newblock In \emph{Proceedings of the IEEE conference on computer vision and pattern recognition}, pages 7291--7299, 2017.

\bibitem[Ci et~al.(2023)Ci, Wang, Chen, Tang, Bai, Zhu, Zhao, Yu, Qi, and Ouyang]{UniHCP}
Yuanzheng Ci, Yizhou Wang, Meilin Chen, Shixiang Tang, Lei Bai, Feng Zhu, Rui Zhao, Fengwei Yu, Donglian Qi, and Wanli Ouyang.
\newblock Unihcp: A unified model for human-centric perceptions.
\newblock \emph{2023 IEEE/CVF Conference on Computer Vision and Pattern Recognition (CVPR)}, pages 17840--17852, 2023.

\bibitem[Dosovitskiy(2020)]{ViT}
Alexey Dosovitskiy.
\newblock An image is worth 16x16 words: Transformers for image recognition at scale.
\newblock \emph{arXiv preprint arXiv:2010.11929}, 2020.

\bibitem[Geng et~al.(2021)Geng, Sun, Xiao, Zhang, and Wang]{DEKR}
Zigang Geng, Ke Sun, Bin Xiao, Zhaoxiang Zhang, and Jingdong Wang.
\newblock Bottom-up human pose estimation via disentangled keypoint regression.
\newblock In \emph{Proceedings of the IEEE/CVF Conference on Computer Vision and Pattern Recognition (CVPR)}, pages 14676--14686, 2021.

\bibitem[Ghasemzadeh et~al.(2021)Ghasemzadeh, Zandycke, Istasse, Sayez, Moshtaghpour, and Vleeschouwer]{DeepSortLab}
Seyed~Abolfazl Ghasemzadeh, Gabriel~Van Zandycke, Maxime Istasse, Niels Sayez, Amirafshar Moshtaghpour, and Christophe~De Vleeschouwer.
\newblock Deepsportlab: a unified framework for ball detection, player instance segmentation and pose estimation in team sports scenes.
\newblock \emph{ArXiv}, abs/2112.00627, 2021.

\bibitem[Gong et~al.(2018)Gong, Liang, Li, Chen, Yang, and Lin]{CIHP}
Ke Gong, Xiaodan Liang, Yicheng Li, Yimin Chen, Ming Yang, and Liang Lin.
\newblock Instance-level human parsing via part grouping network.
\newblock In \emph{Proceedings of the European conference on computer vision (ECCV)}, pages 770--785, 2018.

\bibitem[Gu et~al.(2023)Gu, Chen, and Yao]{Calibration}
Kerui Gu, Rongyu Chen, and Angela Yao.
\newblock On the calibration of human pose estimation.
\newblock \emph{arXiv preprint arXiv:2311.17105}, 2023.

\bibitem[Jin et~al.(2024)Jin, Li, Li, Liu, Qian, and Luo]{HQNet}
Sheng Jin, Shuhuai Li, Tong Li, Wentao Liu, Chen Qian, and Ping Luo.
\newblock You only learn one query: learning unified human query for single-stage multi-person multi-task human-centric perception.
\newblock In \emph{European Conference on Computer Vision}, pages 126--146. Springer, 2024.

\bibitem[Khirodkar et~al.(2021)Khirodkar, Chari, Agrawal, and Tyagi]{MIPNet}
Rawal Khirodkar, Visesh Chari, Amit Agrawal, and Ambrish Tyagi.
\newblock Multi-instance pose networks: Rethinking top-down pose estimation.
\newblock \emph{2021 IEEE/CVF International Conference on Computer Vision (ICCV)}, pages 3102--3111, 2021.

\bibitem[Khirodkar et~al.(2024)Khirodkar, Bagautdinov, Martinez, Su, James, Selednik, Anderson, and Saito]{Sapiens}
Rawal Khirodkar, Timur Bagautdinov, Julieta Martinez, Zhaoen Su, Austin James, Peter Selednik, Stuart Anderson, and Shunsuke Saito.
\newblock Sapiens: Foundation for human vision models.
\newblock In \emph{European Conference on Computer Vision}, 2024.

\bibitem[Li et~al.(2018)Li, Wang, Zhu, Mao, Fang, and Lu]{CrowdPose}
Jiefeng Li, Can Wang, Hao Zhu, Yihuan Mao, Hao-Shu Fang, and Cewu Lu.
\newblock Crowdpose: Efficient crowded scenes pose estimation and a new benchmark.
\newblock \emph{arXiv preprint arXiv:1812.00324}, 2018.

\bibitem[Lin et~al.(2014)Lin, Maire, Belongie, Hays, Perona, Ramanan, Doll{\'a}r, and Zitnick]{COCO}
Tsung-Yi Lin, Michael Maire, Serge~J. Belongie, James Hays, Pietro Perona, Deva Ramanan, Piotr Doll{\'a}r, and C.~Lawrence Zitnick.
\newblock Microsoft coco: Common objects in context.
\newblock In \emph{European Conference on Computer Vision}, 2014.

\bibitem[Ling et~al.(2022)Ling, Huang, and Hur]{HumanPaste}
Evan Ling, De-Kai Huang, and Minhoe Hur.
\newblock Humans need not label more humans: Occlusion copy \& paste for occluded human instance segmentation.
\newblock In \emph{British Machine Vision Conference}, 2022.

\bibitem[Liu et~al.(2021)Liu, Lin, Cao, Hu, Wei, Zhang, Lin, and Guo]{SWIN}
Ze Liu, Yutong Lin, Yue Cao, Han Hu, Yixuan Wei, Zheng Zhang, Stephen Lin, and Baining Guo.
\newblock Swin transformer: Hierarchical vision transformer using shifted windows.
\newblock In \emph{Proceedings of the IEEE/CVF International Conference on Computer Vision}, pages 10012--10022, 2021.

\bibitem[Liu et~al.(2022)Liu, Mao, Wu, Feichtenhofer, Darrell, and Xie]{ConvNeXt}
Zhuang Liu, Hanzi Mao, Chao-Yuan Wu, Christoph Feichtenhofer, Trevor Darrell, and Saining Xie.
\newblock A convnet for the 2020s.
\newblock \emph{Proceedings of the IEEE/CVF Conference on Computer Vision and Pattern Recognition (CVPR)}, 2022.

\bibitem[Lyu et~al.(2022)Lyu, Zhang, Huang, Zhou, Wang, Liu, Zhang, and Chen]{RTMDet}
Chengqi Lyu, Wenwei Zhang, Haian Huang, Yue Zhou, Yudong Wang, Yanyi Liu, Shilong Zhang, and Kai Chen.
\newblock Rtmdet: An empirical study of designing real-time object detectors.
\newblock \emph{ArXiv}, abs/2212.07784, 2022.

\bibitem[Papandreou et~al.(2017)Papandreou, Zhu, Kanazawa, Toshev, Tompson, Bregler, and Murphy]{PoseNMS}
George Papandreou, Tyler~Lixuan Zhu, Nori Kanazawa, Alexander Toshev, Jonathan Tompson, Christoph Bregler, and Kevin~P. Murphy.
\newblock Towards accurate multi-person pose estimation in the wild.
\newblock \emph{2017 IEEE Conference on Computer Vision and Pattern Recognition (CVPR)}, pages 3711--3719, 2017.

\bibitem[Papandreou et~al.(2018)Papandreou, Zhu, Chen, Gidaris, Tompson, and Murphy]{BottomUpSeg}
George Papandreou, Tyler~Lixuan Zhu, Liang-Chieh Chen, Spyros Gidaris, Jonathan Tompson, and Kevin~P. Murphy.
\newblock Personlab: Person pose estimation and instance segmentation with a bottom-up, part-based, geometric embedding model.
\newblock In \emph{European Conference on Computer Vision}, 2018.

\bibitem[Ravi et~al.(2024)Ravi, Gabeur, Hu, Hu, Ryali, Ma, Khedr, R{\"a}dle, Rolland, Gustafson, et~al.]{SAM2}
Nikhila Ravi, Valentin Gabeur, Yuan-Ting Hu, Ronghang Hu, Chaitanya Ryali, Tengyu Ma, Haitham Khedr, Roman R{\"a}dle, Chloe Rolland, Laura Gustafson, et~al.
\newblock Sam 2: Segment anything in images and videos.
\newblock \emph{arXiv preprint arXiv:2408.00714}, 2024.

\bibitem[Rukhovich et~al.(2021)Rukhovich, Sofiiuk, Galeev, Barinova, and Konushin]{IterDet}
Danila Rukhovich, Konstantin Sofiiuk, Danil Galeev, Olga Barinova, and Anton Konushin.
\newblock Iterdet: iterative scheme for object detection in crowded environments.
\newblock In \emph{Structural, syntactic, and statistical pattern recognition: Joint IAPR international workshops, s+ SSPR 2020, padua, Italy, January 21--22, 2021, proceedings}, pages 344--354. Springer, 2021.

\bibitem[Shao et~al.(2018)Shao, Zhao, Li, Xiao, Yu, Zhang, and Sun]{CrowdHuman}
Shuai Shao, Zijian Zhao, Boxun Li, Tete Xiao, Gang Yu, Xiangyu Zhang, and Jian Sun.
\newblock Crowdhuman: A benchmark for detecting human in a crowd.
\newblock \emph{ArXiv}, abs/1805.00123, 2018.

\bibitem[Shi et~al.(2022)Shi, Wei, Li, Ren, and Tan]{PETR}
Dahu Shi, Xing Wei, Liangqi Li, Ye Ren, and Wenming Tan.
\newblock End-to-end multi-person pose estimation with transformers.
\newblock In \emph{Proceedings of the IEEE/CVF Conference on Computer Vision and Pattern Recognition}, pages 11069--11078, 2022.

\bibitem[Stoffl et~al.(2021)Stoffl, Vidal, and Mathis]{POET}
Lucas Stoffl, Maxime Vidal, and Alexander Mathis.
\newblock End-to-end trainable multi-instance pose estimation with transformers.
\newblock \emph{arXiv preprint arXiv:2103.12115}, 2021.

\bibitem[Sun et~al.(2019)Sun, Xiao, Liu, and Wang]{HRNet}
Ke Sun, Bin Xiao, Dong Liu, and Jingdong Wang.
\newblock Deep high-resolution representation learning for human pose estimation.
\newblock In \emph{Proceedings of the IEEE conference on computer vision and pattern recognition}, pages 5693--5703, 2019.

\bibitem[Sun et~al.(2024)Sun, Gu, Wang, Yang, and Yao]{RethinkingVisibility}
Pengzhan Sun, Kerui Gu, Yunsong Wang, Linlin Yang, and Angela Yao.
\newblock Rethinking visibility in human pose estimation: Occluded pose reasoning via transformers.
\newblock \emph{2024 IEEE/CVF Winter Conference on Applications of Computer Vision (WACV)}, pages 5891--5900, 2024.

\bibitem[Tripathi et~al.(2017)Tripathi, Collins, Brown, and Belongie]{pose2instance}
Subarna Tripathi, Maxwell~D. Collins, Matthew~A. Brown, and Serge~J. Belongie.
\newblock Pose2instance: Harnessing keypoints for person instance segmentation.
\newblock \emph{ArXiv}, abs/1704.01152, 2017.

\bibitem[Wang and Zhang(2022)]{CID}
Dongkai Wang and Shiliang Zhang.
\newblock Contextual instance decoupling for robust multi-person pose estimation.
\newblock \emph{2022 IEEE/CVF Conference on Computer Vision and Pattern Recognition (CVPR)}, pages 11050--11058, 2022.

\bibitem[Wang et~al.(2023{\natexlab{a}})Wang, Dai, Chen, Huang, Li, Zhu, Hu, Lu, Lu, Li, et~al.]{InternImage}
Wenhai Wang, Jifeng Dai, Zhe Chen, Zhenhang Huang, Zhiqi Li, Xizhou Zhu, Xiaowei Hu, Tong Lu, Lewei Lu, Hongsheng Li, et~al.
\newblock Internimage: Exploring large-scale vision foundation models with deformable convolutions.
\newblock In \emph{Proceedings of the IEEE/CVF conference on computer vision and pattern recognition}, pages 14408--14419, 2023{\natexlab{a}}.

\bibitem[Wang et~al.(2023{\natexlab{b}})Wang, Wu, Tang, He, Guo, Zhu, Bai, Zhao, Wu, He, and Ouyang]{HULK}
Yizhou Wang, Yixuan Wu, Shixiang Tang, Weizhen He, Xun Guo, Feng Zhu, Lei Bai, Rui Zhao, Jian Wu, Tong He, and Wanli Ouyang.
\newblock Hulk: A universal knowledge translator for human-centric tasks.
\newblock \emph{ArXiv}, abs/2312.01697, 2023{\natexlab{b}}.

\bibitem[Wu et~al.(2017)Wu, Zheng, Zhao, Li, Yan, Liang, Wang, Zhou, Lin, Fu, Wang, and Wang]{AIC}
Jiahong Wu, He Zheng, Bo Zhao, Yixin Li, Baoming Yan, Rui Liang, Wenjia Wang, Shipei Zhou, Guosen Lin, Yanwei Fu, Yizhou Wang, and Yonggang Wang.
\newblock Ai challenger : A large-scale dataset for going deeper in image understanding.
\newblock \emph{ArXiv}, abs/1711.06475, 2017.

\bibitem[Xiao et~al.(2022)Xiao, Wang, Yu, Su, Jin, Song, Yan, and Zhao]{AdaptivePose++}
Yabo Xiao, Xiaojuan Wang, Dongdong Yu, Kai Su, Lei Jin, Mei Song, Shuicheng Yan, and Jian Zhao.
\newblock Adaptivepose++: A powerful single-stage network for multi-person pose regression.
\newblock \emph{arXiv preprint arXiv:2210.04014}, 2022.

\bibitem[Xu et~al.(2022)Xu, Zhang, Zhang, and Tao]{ViTPose}
Yufei Xu, Jing Zhang, Qiming Zhang, and Dacheng Tao.
\newblock Vi{TP}ose: Simple vision transformer baselines for human pose estimation.
\newblock In \emph{Advances in Neural Information Processing Systems}, 2022.

\bibitem[Yuan et~al.(2021)Yuan, Fu, Huang, Lin, Zhang, Chen, and Wang]{HRFormer}
Yuhui Yuan, Rao Fu, Lang Huang, Weihong Lin, Chao Zhang, Xilin Chen, and Jingdong Wang.
\newblock Hrformer: High-resolution vision transformer for dense predict.
\newblock \emph{Advances in neural information processing systems}, 34:\penalty0 7281--7293, 2021.

\bibitem[Zhang et~al.(2019)Zhang, Li, Dong, Rosin, Cai, Han, Yang, Huang, and Hu]{pose2seg}
Song-Hai Zhang, Ruilong Li, Xin Dong, Paul Rosin, Zixi Cai, Xi Han, Dingcheng Yang, Haozhi Huang, and Shi-Min Hu.
\newblock Pose2seg: Detection free human instance segmentation.
\newblock In \emph{Proceedings of the IEEE/CVF conference on computer vision and pattern recognition}, pages 889--898, 2019.

\bibitem[Zhou and He(2020)]{PoSeg}
Desen Zhou and Qian He.
\newblock Poseg: Pose-aware refinement network for human instance segmentation.
\newblock \emph{IEEE Access}, 8:\penalty0 15007--15016, 2020.

\bibitem[Zhou et~al.(2023)Zhou, Stoffl, Mathis, and Mathis]{BUCTD}
Mu Zhou, Lucas Stoffl, Mackenzie~W. Mathis, and Alexander Mathis.
\newblock Rethinking pose estimation in crowds: overcoming the detection information bottleneck and ambiguity.
\newblock \emph{2023 IEEE/CVF International Conference on Computer Vision (ICCV)}, pages 14643--14653, 2023.

\bibitem[Zong et~al.(2023)Zong, Song, and Liu]{coDETR}
Zhuofan Zong, Guanglu Song, and Yu Liu.
\newblock Detrs with collaborative hybrid assignments training.
\newblock In \emph{Proceedings of the IEEE/CVF international conference on computer vision}, pages 6748--6758, 2023.

\end{thebibliography}
